\documentclass[pdflatex,sn-mathphys-num]{sn-jnl}

\usepackage{amsthm}%
\usepackage{mathrsfs}%
\usepackage[title]{appendix}%
\usepackage{manyfoot}%
\usepackage{booktabs}%
\usepackage{algorithmicx}%
\usepackage{listings}%

\usepackage{graphicx}
\usepackage{array}
\usepackage{multirow}
\usepackage{rotating}
\usepackage{booktabs} 
\usepackage{multirow}
\usepackage{pgffor} 

\usepackage{subfig}

\usepackage{adjustbox}

\usepackage{hyperref}
\usepackage{amsmath,amssymb,amsfonts}

\usepackage{algorithm} 
\usepackage{algpseudocode} 

\usepackage{textcomp}
\usepackage{xcolor}

\usepackage{diagbox} 

\theoremstyle{thmstyleone}%
\theoremstyle{thmstyletwo}%

\theoremstyle{thmstylethree}%

\begin{document}

\title[Race Bias Free Face Aging]{A Race Bias Free Face Aging Model for Reliable Kinship Verification}

\author{\fnm{Ali} \sur{Nazari}}\email{al\_nazari@sbu.ac.ir}
\author{\fnm{Bardiya} \sur{Kariminia}}\email{b.kariminia@mail.sbu.ac.ir}
\author*{\fnm{Mohsen} \sur{Ebrahimi Moghaddam}}\email{m\_moghadam@sbu.ac.ir}

\affil{\orgdiv{Faculty of Computer Science and Engineering}, \orgname{Shahid Beheshti University}, \orgaddress{\street{Velenjak}, \city{Tehran}, \postcode{1983969411}, \state{Tehran}, \country{Iran}}}

\abstract{The age gap in kinship verification addresses the time difference between the photos of the parent and the child. Moreover, their same-age photos are often unavailable, and face aging models are racially biased, which impacts the likeness of photos. Therefore, we propose a face aging GAN model, RA-GAN, consisting of two new modules, RACEpSp and a feature mixer, to produce racially unbiased images. The unbiased synthesized photos are used in kinship verification to investigate the results of verifying same-age parent-child images. The experiments demonstrate that our RA-GAN outperforms SAM-GAN on an average of 13.14\% across all age groups, and CUSP-GAN in the 60+ age group by 9.1\% in terms of racial accuracy. Moreover, RA-GAN can preserve subjects' identities better than SAM-GAN and CUSP-GAN across all age groups. Additionally, we demonstrate that transforming parent and child images from the KinFaceW-I and KinFaceW-II datasets to the same age can enhance the verification accuracy across all age groups. The accuracy increases with our RA-GAN for the kinship relationships of father-son and father-daughter, mother-son, and mother-daughter, which are 5.22, 5.12, 1.63, and 0.41, respectively, on KinFaceW-I. Additionally, the accuracy for the relationships of father-daughter, father-son, and mother-son is 2.9, 0.39, and 1.6 on KinFaceW-II, respectively. The code is available at~\href{https://github.com/bardiya2254kariminia/An-Age-Transformation-whitout-racial-bias-for-Kinship-verification}{Github}
}


\maketitle

\section{Introduction}
The Kinship Verification (KV) algorithm accepts two facial images and determines whether they have a family relationship. One of the main challenges in this area is that facial images have been taken at different times, making it difficult to verify their relationship since they have varying features. This age gap between generations degrades verification accuracy due to the varying age conditions on the faces of parents and children. Some researchers~\cite{wang2018cross} attempted to transform the facial image of the parent into that of a younger one and then verify their kinship. 

Another study investigated KV in childhood and claimed that parents and children share similar characteristics~\cite{oruganti2023deep} at a very young age.
Although these can be fruitful, the wildly varying age gap cannot be completely mitigated, and they did not consider all age ranges. We aim to transform the facial images of parents and children to the same age in various age ranges and investigate the KV accuracy. Our work can testify to which opinion can better mitigate age variation, as we investigate facial images of individuals of all ages, from childhood to adulthood.

Furthermore, since facial images of individuals of all ages are not always available, we utilize age transformation to synthesize images of various ages. 
Moreover, face age transformation methods predominantly suffer from racial bias, such as SAM-GAN~\cite{alaluf2021only}, which impacts the likeness between original and synthesized images. This likeness is significant in KV since we want to verify the same parent and child at various ages. Therefore, any age transformation must include the avoidance of biases, especially those that influence perceptions of likeness, such as race. To address this issue, we propose an age transformation model, RA-GAN, which attempts to remove this bias from the synthesized images.
Additionally, since the datasets of age transformation methods are imbalanced with respect to attributes such as race, they synthesize images with biases. 
Popular facial data sets such as FFHQ~\cite{karras2019style}, LFW~\cite{huang2008labeled}, and CelebA~\cite{liu2015deep}, are racially unbalanced and tend to white people roughly $70\%$, $77.5\%$ and Asian ethnicity, respectively. It is significant that these data sets should be free of race bias, since training the network with those datasets leads to inaccurate representation and incorrect image creation. Therefore, we compiled a dataset free of racial bias.


In simple form, we contribute three parts:
\begin{itemize}
\item A new face aging model, called RA-GAN, has been developed and consists of two novel modules: 1) a Race Encoder (called RACEpSp), which eliminates racial bias, and 2) a Feature Mixture, which fuses age features and racially specific face features. It also preserves subjects' identities better than state-of-the-art algorithms, such as CUSP and SAM-GAN.
\item We also created a new dataset by collecting data from the UTKFace dataset, a race-balanced dataset, and conducted our experiment using this racially fair dataset.
\item We thoroughly portray the influence of the same-age images on kinship verification to increase its accuracy.
\end{itemize} 

The organization of this paper is as follows: Section~\ref{sec:related-work} consists of two subsections: Subsection~\ref {subsec:face-aging} explains the related work in face aging models. Subsection~\ref {subsec:aging-in-kinship} describes those KV studies that used face aging models. Our face aging model is proposed in Section~\ref{sec:proposed-method}. The general schema for using the face aging model in KV is presented in Subsection~\ref {subsec:kinship-ragan}. All experiments are represented in Section~\ref{sec:experiment}. Ultimately, we conclude the paper in Section~\ref{sec:conclusion}.

\section{Related Work}
\label{sec:related-work}
We first investigate studies of aging in kinship verification in~\ref{subsec:aging-in-kinship}, then in generating facial images~\ref{subsec:face-aging}, and finally aging and race in both face-generated images~\ref{subsec:race-in-aging}.

\subsection{Aging in Kinship Verification}
\label{subsec:aging-in-kinship}
\begin{figure*}[ht]
\centering
\subfloat[Hierarchy of Kinship Face Synthesis\label{fig:hierarchy-Kinship-synthesis}]{
\begin{minipage}[c]{0.45\linewidth}
	\centering
	\includegraphics[width=\linewidth]{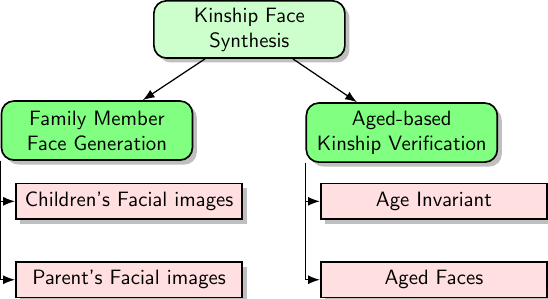}
\end{minipage}	
}
\hfill
\subfloat[Hierarchy of Face Aging\label{fig:hierarchy-face-aging}]{
\begin{minipage}[c]{0.45\linewidth}
	\centering
	\includegraphics[width=\linewidth]{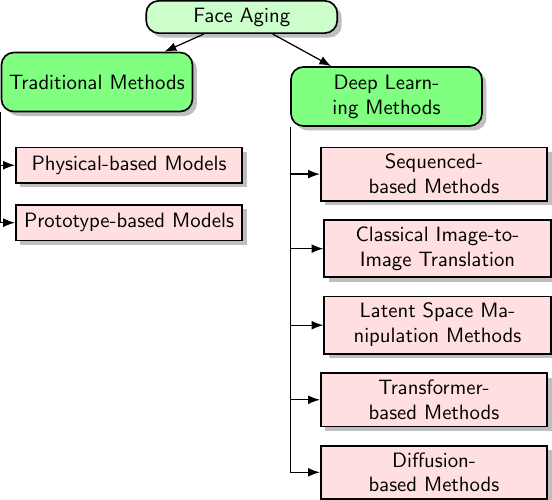}
\end{minipage}
}
\caption{Hierarchy of face aging and kinship face synthesis methods}
\label{fig:hierarchy-both-kinship-face}	
\end{figure*}

Two kinds of research produce facial images from kin-related images, as shown in Fig.~\ref{fig:hierarchy-Kinship-synthesis}. 
First, some researchers attempted to produce a facial image of a family member, divided into two subcategories: child-generated and parent-generated images.
Second, some other researchers aim to increase the accuracy of KV by generating people's facial images at different ages or extracting features regardless of age variations. This track focuses on surmounting the generation gap or the divergence of aged facial images.

The primary focus in the first branch is to generate kin-related facial images. Some researchers' attention has been absorbed by synthesizing children's facial pictures from their parents' facial images, such as KinshipGAN~\cite{ozkan2018kinshipgan}, DNA-Net~\cite{gao2021dna}, AdvKin~\cite{zhang2020advkin}, FamilyGAN~\cite{sinha2020familygan}, GANKIN~\cite{ghatas2020gankin}, ChildNet~\cite{pernuvs2023childnet}, StyleGene~\cite{li2023stylegene}, StyleDiT~\cite{chiu2024styledit}. In contrast, others attempted to synthesize the parent's facial image from the child's facial image, ParentGAN~\cite{emara2024parent}.

Moreover, the primary purpose of the second branch is to enhance the accuracy of kinship verification by utilizing synthesized images to address age variation. The young cross-generational model~\cite{wang2018cross} used a discriminative metric and rejuvenated parental facial images to mitigate identity and age variation. It stated that it is highly probable that a parent's and a child's facial images share common features due to the high similarity of their young images.
The face age transform model~\cite{kim2023robust} employed an auto-encoder structure by injecting age features in the latent space to produce facial images in various ages and then fined-tuned the Inception-ResNet to extract features for KV. Childhood stage model~\cite{oruganti2023deep} was based on the opinion that facial images of parent and their children are akin. Thus, it converted both parent and child images to childhood and then computed the accuracy of the kinship verification on a proposed collected childhood dataset.
Additionally, age-invariant adversarial feature learning~\cite{liu2022age} attempted to separate the identity features from the age features and then compute the verification. 

Furthermore, in~\cite{chandaliya2022childgan} a self-attention module block has been developed for extracting the channel-wise and spatial correlation on the input latent code and trained the model with a generator conditioned on the age. On the contrary, there exists limited training data for children; for instance, FG-NET, CACD, or MORPH are preferred datasets used for aging but lack diversity and scarcity across age groups. Also, the further we dive into the future predictions, the less accurate and more speculative the output we receive. Also, the model can be biased towards certain racial groups such as Indian which is not desirable for the prediction.

However, despite aiming for personalization, the model may not generalize well to underrepresented age groups, races, or facial structures in the training set. This limits fairness and robustness.This effect can  be caused by training the model purely on the used dataset which is not ethnic distribution balanced dataset.Moreover, the age group categorized in discrete classes and is not continuous.   

So, if the generator's domain is biased towards a certain race or identity because of training separately on a racially biased dataset, no more speculation and editing could be done to fix the generator's prediction manifold. The more enhancement we can make is to find the best possible encoder to approximate the proper mapping on the generator's distribution.

\subsection{Age Transformation Studies in Face}
\label{subsec:face-aging}
In the progression and regression of the age of the face, an image of older or younger age is constructed using an individual face image related to the specified age~\cite{grimmer2021deep}. Methods in this area can be categorized as early traditional or deep learning methods as shown with their subcategories in Fig.~\ref{fig:hierarchy-face-aging}. 

Traditional aging models can be categorized into physical-based~\cite{ramanathan2008modeling} and prototype-based models~\cite{tang2017personalized}. Physical-based models attempt to model an individual's facial structure, such as muscles, skin, and brain skull. Everyone has a specific model, which is not generalizable to others besides its heavy computation. Data sets of this approach have required the inclusion of everyone's face images over the years. This approach has some negative points, such as lack of generalization, heavy computation, and a data set of all subjects' different face images over the years~\cite{grimmer2021deep}.

In addition, prototype-based or exemplar methods intend to find various representatives for each age group. These representatives are determined by aggregating facial components, illumination, and texture information in each age group~\cite{georgopoulos2018modeling}. To create the synthesized younger or older image, these representatives are combined with an individual's face information. The identity of the synthesized images is not the same as the original identity, and the images have ghost artifacts~\cite{grimmer2021deep}.

In addition, deep learning methods, popular due to their high accuracy, scalability, and end-to-end learning, have been used in face aging. Their advantages in this realm include identity preservation and high image fidelity, and they do not have the drawbacks of traditional methods. We divide face aging using deep learning into five categories: sequence-based, classical image-to-image translation, latent space manipulation, transformer-based, and diffusion-based methods~\cite{guo2024age}.

\subsubsection{Sequenced-based Methods}
The human age has gradually changed. Age transformation from one group to the nearby age group can be smooth and continuous because nearby age groups are close. Thus, these consecutive transformations can be modeled as a sequence by Recurrent Neural Networks (RNN)~\cite{wang2016recurrent}, Markov Process~\cite{suo2012concatenational, suo2009compositional} or Deep Restricted Boltzmann Machines (RBM)~\cite{nhan2016longitudinal, nhan2017temporal}. In the graph evolution aging model~\cite{suo2012concatenational}, changes in each face component were modeled separately in short-term and long-term views. An aging function in the short-term view for each age group and a Markov Process in the long-term view were leveraged. In the recurrent face aging~\cite{wang2016recurrent}, an RNN with two-layer Gated Recurrent Units (GRU) was adopted between shared eigenfaces of two neighboring groups. In the longitudinal face modeling via TRBM~\cite{nhan2016longitudinal}, the face structure and aging features were modeled using RBMs. High-level features were transferred from hidden units between RBMs of various age groups. Other enhancements, such as wrinkles and shape, were adopted.

\subsubsection{Classical Image-to-Image Translation Methods}
Image-to-image translation techniques aim to change image samples of a source domain to a target domain by injecting essential features from the source into the target domain. This approach retains the image's identity and changes the image to a proper sample in the target domain according to the style or appearance of the target domain.

Conditional GAN~\cite{isola2017image} was applied by training the model with paired data sets and considering contextual information. Cycle-GAN~\cite{zhu2017unpaired} has enhanced by using a cycle consistency loss, an $l_1norm$ between the source and reconstructed output collections. Cycle-GAN pioneered in this field as it did not require samples to be in pairs, but it also required retraining every mapping between two collections. StarGAN~\cite{choi2018stargan} tackled the problem of training multiple models for different source-target mapping by adding a mask vector to the domain label and introducing a domain classification loss, $\mathcal{L}_{cls}$, to ensure proper transformation. However, these methods were inefficient in face aging translation since the general face information related to the specific age of samples changed during translation~\cite{li2023gfam}.

In face aging, the Age-cGAN model~\cite{antipov2017face} used the age vector as a conditional parameter to the generator and an identity-preserving module checking the similarity of the FaceNet features between the original and targeted face images. 

Re-aging GAN~\cite{makhmudkhujaev2021re} employed an age modulator module receiving the target age and encoded features. Then, the output of the age modulator was spread over all decoder layers. Adversarial, reconstruction, and cycle-consistency losses were leveraged to preserve identity and obtain fewer artifacts and natural transformation.

\subsubsection{Latent Space Manipulation Methods}
With the emergence of well-trained GAN generators, researchers have attempted to take full advantage of them and explore meaningful changes in their latent space, either by learning the mapping of the desired attribute or by traversing it. Most latent space manipulation methods assume that the latent space is linear, traversal, and controllable, try to find the corresponding point of the input images in the GAN latent space first introduced in~\cite{zhu2016generative} called GAN Inversion, and then edit the found latent representation to generate the desired image. 

This method has some challenges; for example, finding a unique GAN latent representation for the real images is difficult, especially when we want to synthesize an image of the desired age and the input image has not been in the training set.
The most widely used generator of facial images is StyleGAN~\cite{karras2019style}, presenting high-quality images with diversity. Its latent space~$W$ is more disentangled than its predecessors' latent space $Z$. The adaptive instance normalization has also been used in StyleGAN and redesign in StyleGAN-v2 to extract styles from latent representations. This model depicts a better semantic editing on features such as pose, expression, style of images and ages.

Moreover, StyleFlow~\cite{abdal2021styleflow} leaned the nonlinear path in the latent space and aimed to make attribute-conditioned edits using normalizing flows. It applied a conditional normalization flow mapping between latent and attribute vectors, enabling the editing process to become invertible and smooth due to the disentanglement of the latent vector representation and the attribute vector representation. This capability allowed editing different image features like jawline, hair color, luminance, skin color, and age.

In addition, SAM-GAN~\cite{alaluf2021only} has attempted to learn the nonlinear path in the latent space of StyleGAN, preserve the identity, and change the age. It designed an age encoder alongside the pixel2style2pixel~(pSp) encoder~\cite{richardson2021encoding} to enable the injection of the age attribute into the image's latent representation.

One point about models like SAM-GAN~\cite{alaluf2021only} or StyleFlow~\cite{abdal2021styleflow} is that they cannot change the manifold or domain of the generator owing to freezing its weights and not being trained. Additionally, their conversion often produced undesirable effects on the output images, known as the trade-off between reconstruction and editing quality~\cite{katsumata2024revisiting}.

Furthermore, ChildGAN~\cite{chandaliya2022childgan} was introduced for finding missing children and preserving the racial features of children's images. It was composed of Encoder, Generator, and Attention Discriminator. Its self-attention block could extract the channel-wise and spatial correlation on the output of its encoder. Age and gender information were coded as one-hot vectors and combined with the latent space to feed the generator. For the lack of children's data and racial diversity, children's images from various sources and the internet are collected. Also, its model can be biased towards certain racial groups, such as Indians, which is not desirable. Moreover, the age groups were categorized in discrete classes and not continuous.

\subsubsection{Transformer-based Methods}
TransGAN~\cite{jiang2021transgan} displayed the first pure transformer-based GAN without any convolution layers, outperforming convolution-based GAN like StyleGAN-V2, especially in generating images with high-resolution, high fidelity, and texture details. Its grid self-attention can alleviate the memory bottleneck and allow scale-up in generating high-resolution images. HiT~\cite{zhao2021improved} and StyleSwim~\cite{zhang2022styleswin} were other transformer-based GANs to generate high-resolution images. HiT leveraged the multi-axis blocked self-attention to mix local and global attention and dropped self-attention in high-resolution stages. StyleSwim utilized the Swin transformer in the generator with a Style-based architecture and gained a greater receptive field with double attention. It also leveraged wavelets in the discriminator to tackle the blocking artifact.

\subsubsection{Diffusion-based Methods}
Diffusion models~\cite{ho2020denoising}, owing to their promising results in modeling continuous data, have been utilized in image-to-image translation, such as Palette~\cite{saharia2022palette}, a framework using diffusion models instead of GANs. It gradually added small amounts of Gaussian noise to images during training and eliminated the noise during inference to generate realistic images, producing high-quality outputs. However, it required an extensive computational resource and was slower in inference time. FADING~\cite{chen2023face} leveraged a large-scale language-image diffusion model to surmount face aging and also the large age gap.



In conclusion, to our knowledge, all studies considering age variation in KV did not simultaneously cope with race. Additionally, they did not thoroughly investigate the effect of all possible age ranges on KV.

\section{Proposed Method}
\label{sec:proposed-method}
\begin{figure*}[h]
\begin{center}
\includegraphics[width=\textwidth]{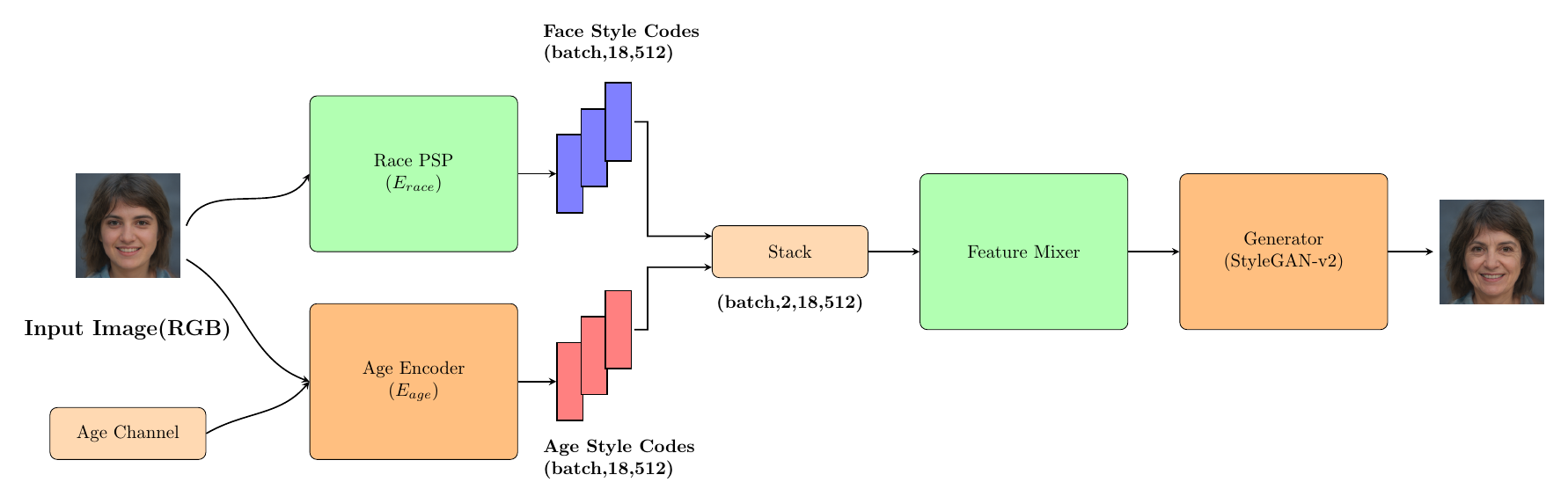}
\caption{The General Diagram of the Proposed Method. The green entities are trainable, and the orange ones are frozen.}
\label{fig:propose-method}
\end{center}
\end{figure*}

To realize the idea, a method is proposed and generally depicted in Fig.~\ref{fig:propose-method}
We refer to our model as the RA-GAN (RaceAgingGAN) framework.

\subsection{Overview}
Face aging takes the input image $x_{in}$ with its age $age_{in}$ and generates the output image $x_{out}$ at the desired target age $age_t$. Our concern is preserving the synthesized image's identity and, notably, race according to (\ref{eq:face-aging-formulation}).
\begin{align}
x_{out}  &= \text{RA-GAN} \left( x_{in} ,\ {age}_{t} \right) \notag \\
s.t \quad & \text{identity}(x_{\text{out}}) = \text{identity}(x_{\text{in}}) \notag \\
& \text{race}(x_{out}) = \text{race}(x_{in})  \notag \\
& \text{age}(x_{out})= age_{t}
\label{eq:face-aging-formulation}
\end{align}

Our face aging, based on the image-to-image translation technique and latent code manipulation, consists of the pre-trained StyleGAN-V2 generator, the pixel2style2pixel~(pSp)~\cite{richardson2021encoding} age encoder, a modified pSp module we call RACEpSp to augment race information, and a feature mixer module proposed to mix and find the best possible combination from the features of age and race information.

The lack of a racially fair dataset with various samples and poses makes face aging challenging. With the assistance of RACEpSp, we can negate the effect of generations of racially biased face images.

\subsection{Architecture of RACEpSp}
Various facial images with varying poses affect the training procedure because the pre-trained StyleGAN generator tends to map images to a frontal face pose. Consequently, images with different face poses won't be mapped correctly in the generator's manifold, resulting in different identities and altered race.

To address the issue of race bias, we utilize a pre-trained ResNet34 model on the Fairface dataset, which comprises over 100,000 samples with diverse face poses and is free from racial biases. To retrieve the lost data in the pyramid network of RACEpSp, which helps the model map images better on the generator's manifold. 

The latent features of the synthesized facial images represent a manifold in the latent space. The more unbiased the data, the more precise the manifold is. Biased data can also affect the model's generalization and the manifold's form since the synthesized images from biased data are invalid and represent another person. The RACEpSp module is proposed to include race features in the final features to make them racially unbiased. 

\begin{figure*}[htp]
\begin{center}
\includegraphics[width=\textwidth]{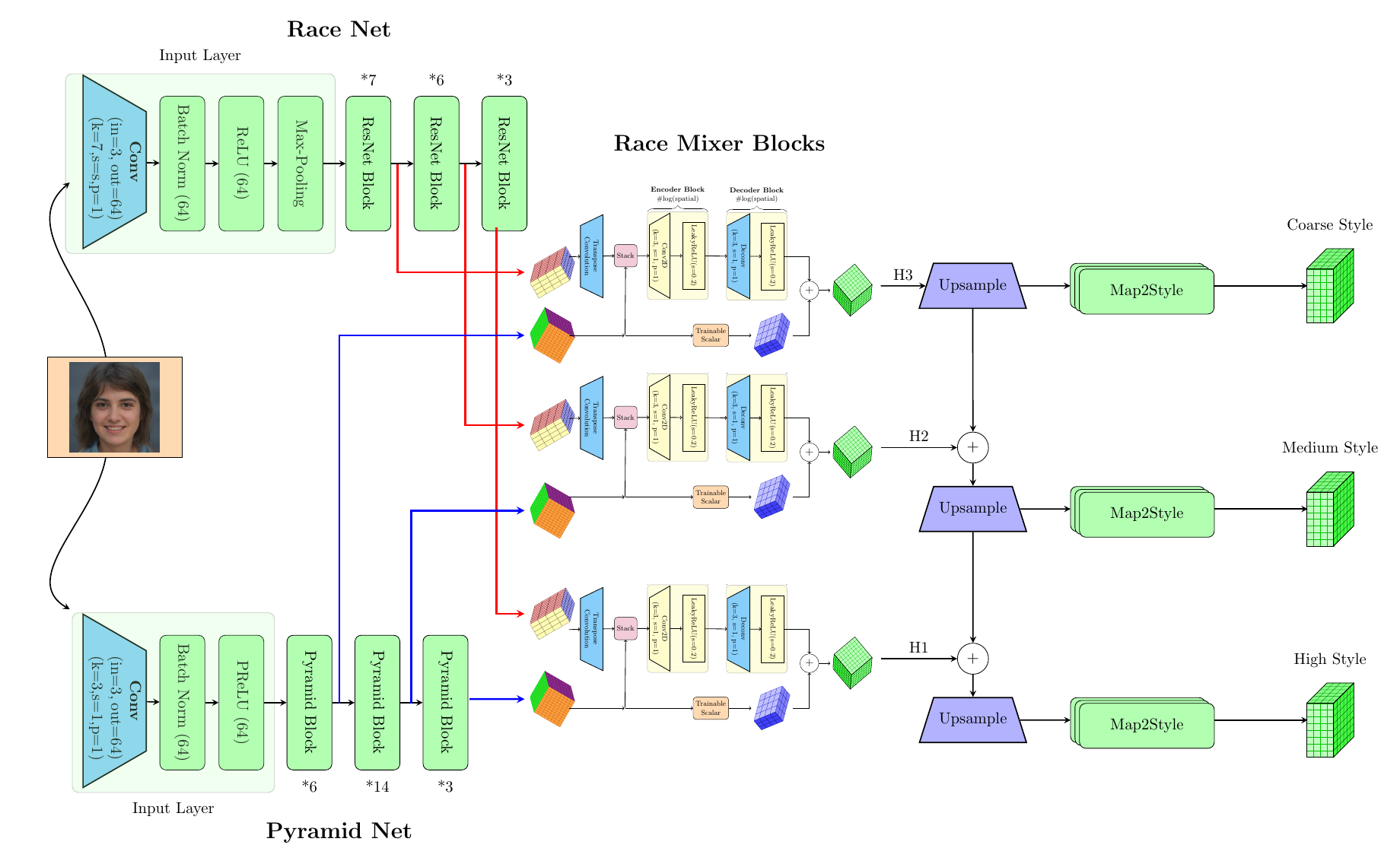}
\caption{The diagram of the RACEpSp module taking three feature vectors of face and race, mixing them up with a transpose convolution layer and an auto-encoder architecture network, and augmenting them with the scaled facial information. The RaceNet and PyramidNet sub-neural networks are frozen. Other elements are trainable.}
\label{fig:RacepSp}
\end{center}
\end{figure*}

The RACEpSp module represented in Fig.~\ref{fig:RacepSp} takes the input image $x_{in}$ with three RGB channels and feeds it into two sub-neural networks: RaceNet and PyramidNet.

\subsubsection{RaceNet}
The sub-neural network RaceNet, which consists of residual blocks, is trained on various face images with diverse poses and races to mitigate bias related to race and facial poses. 
\subsubsection{PyramidNet}
We leverage the pre-trained PyramidNet.
The sub-neural network, PyramidNet, consists of pyramid blocks that produce face features with different resolutions in the feature space.

The features of the 1$^{st}$, 2$^{nd}$, and 3$^{rd}$ pyramid blocks from PyramidNet and features of 7, 13, and 15 residual blocks from RaceNet, respectively, are fed to their corresponding race mixer blocks. Each RaceMixer accepts a proper amount of both racial and facial information. Consequently, three feature vectors~$H_i$ are gained from each RaceMixer block. Each $H_i$ is augmented by the upsampled $H_{i-1}$ from the previous RaceMixer as we attempt to compensate for information loss in pyramid convolutions. The output of each up-sampling layer will then be fed into various Map2Style blocks, introduced in the pSp framework, which create different style codes.

Our idea for proposing RaceMixer is to fuse racial and facial information. If a linear layer or their concatenation is leveraged to mix them, the dimension of the fused vector will be huge. We know that the parameters of a set of convolutions are much lower than those of a linear layer with the same input size. In convolution, one function is impacted and modified by another, known as a kernel function. Thus, the convolutional layers are employed to transform the face features with race features. Moreover, we want to transform the blended features into the right space of facial information. These reasons lead us to the architecture of the autoencoder, which incorporates convolutional layers. In addition, the dimensions of facial and racial features from various levels of the network differ; before feeding them to the autoencoder, a transposed convolution is employed to equalize their dimensions.

Since the facial information is vital and must not be lost, the second pipeline with a scalar layer is designed. The scalar layer learns this precise amount of face information. Additionally, as the features of deep and shallow layers differ in both the sub-neural networks PyramidNet and RaceNet, separate RaceMixer blocks are considered for each level.

\begin{figure*}[htp]
\begin{center}
\includegraphics[width=\textwidth]{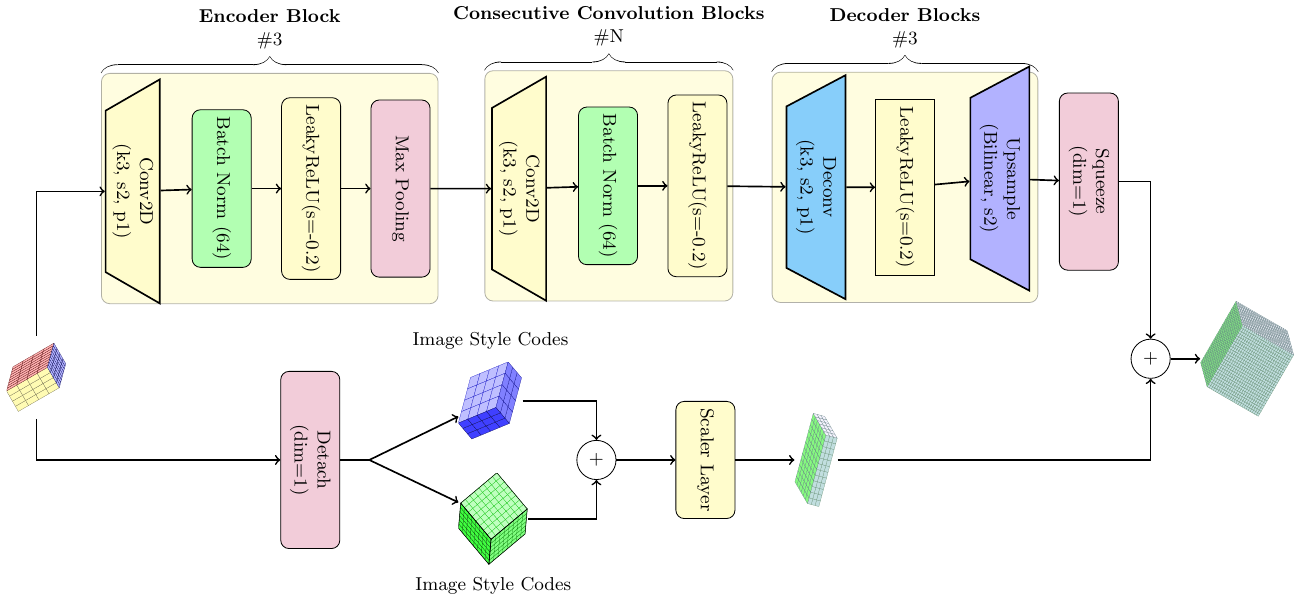}
\caption{The standard schema of the Feature mixer for mixing the proper amount  of data's from different style codes}
\label{fig:feature-mixer}
\end{center}
\end{figure*} 

\subsection{Architecture of AgeEncoder and Generator}
The architectures of our AgeEncoder and generator are similar to those of SAM-GAN~\cite{alaluf2021only} and StyleGAN-V2, respectively.

\subsection{Training CNN networks}

First, due to the effectiveness of normalizing colored face images in training fast and stably, they are normalized~\cite{yousif2023enhancing,albert2023comparison}.

A desired target age is randomly generated from a uniform distribution between 10 and 80 years old according to~(\ref{eq:uniform-age}).
\begin{equation}
\alpha_t \sim  \mathcal{U}(10,80)
\label{eq:uniform-age}
\end{equation}
This sampled age is used as the fourth channel, alongside the RGB channels, of the facial image~(\ref{eq:four-channel}). 
\begin{equation}
x_{age}= x || \alpha_t
\label{eq:four-channel}
\end{equation}
This $x_{age}$ is fed to the AgeEncoder module~$E_{age}(.)$ producing age-style codes~$S_{age}$~(\ref{eq:age-style-codes})
\begin{equation}
S_{age} = E_{age}(x_{age})\in \mathbb{R}^{(18,512)}
\label{eq:age-style-codes}
\end{equation}

The age-style codes $S_{age}$ are the desired latent codes responsible for changing the mapping location of the $\mathcal{W+}$ manifold on the StyleGAN-V2 generator.

Furthermore, the image $x$ is fed to the RACEpSp module $E_{race}(.)$ to receive the racially face style codes on the $\mathcal{W+}$ manifold of the StyleGAN-V2 generator. 
\begin{equation}
S_{face} = E_{race}(x)\in \mathbb{R}^{(18,512)}
\label{eq:face-style-codes}
\end{equation}

In the final phase, we stack up the age and racially face style codes and feed them to the feature mixer module depicted in Fig~\ref{fig:feature-mixer} to combine face style codes with not only the same age style code but also more correlated data, which make the latent features to be mapped on the precise location on the manifold. 

We denote the feature mixer module as $F(.)$ and its output as $F_{mix}$ according to~(\ref{eq:feature-mixture}).
\begin{equation}
F_{mix} = F(stack(S_{age},S_{face})) \in \mathbb{R}^{(18,512)}
\label{eq:feature-mixture}
\end{equation}

In the end, we feed the $F_{mix}$ to the StyleGAN-V2 generator $G(.)$ and receive the synthesized images according to~(\ref{eq:stylegan-generator})
\begin{equation}
x'= G(F_{mix}) \in \mathbb{R}^{(3,256,256)}
\label{eq:stylegan-generator}
\end{equation}
The entire process of the framework is summarized in~(\ref{eq:rage})
\begin{align}
x'& = \text{RA-GAN}(x,\alpha_t) \notag \\
& = G(F(stack(E_{age}(x || \alpha_t) , E_{race}(x) )))
\label{eq:rage}
\end{align}

\subsection{Loss Functions}
Our framework has been trained using a combination of some loss functions.

First, we utilize the losses $\mathcal{L}_2$ and $\mathcal{L}_{id}$ to enable our model to learn pixel and perceptual patch similarities, respectively.
\begin{equation}
\mathcal{L}_2(x, x^\prime ) =||x-x^\prime||_2 = ||x-\text{RA-GAN}(x, \alpha_{t})||_2
\label{eq:L-2}
\end{equation}
\begin{equation}
\mathcal{L}_{id}(x, x^\prime) = ||F_{ArcFace}(x) - F_{ArcFace}\left(x^\prime\right)||_2
\label{eq:L-pips}
\end{equation}

Here, $F_{ArcFace}(.)$ denotes a feature extractor of the pre-trained ResNet-50 model, its last CNN layer, used to extract meaningful features for the loss function $\mathcal{L}_{id}$.

\begin{equation}
\mathcal{L}_{Aging}(x, x^\prime) = ||F_{DEX}(x) - F_{DEX}\left(x^\prime\right)||_2
\label{eq:L-id}
\end{equation}

$F_{DEX}(.)$ denotes a feature vector of the last CNN layer before the fully connected layer in the VGG16 model pre-trained for the age estimation~\cite{rothe2015dex}.

We employ a regularization loss known as the $w_{norm}$ loss~(\ref{eq:L-w-norm}) to remove ghosting artifacts and achieve a high-quality image.
\begin{equation} 
\mathcal{L}_{w_{norm}}(F_{mix}) = \frac{\sum \|\text{F}_{mix} \|_2}{\text{batch-size}}
\label{eq:L-w-norm}
\end{equation}

To satisfy the race condition, we use 
\begin{equation}
\mathcal{L}_{race}(x , x') = ||F_{racenet}(x) - F_{racenet}(x')||_2
\label{eq:L-race}
\end{equation}

Here, $F_{racenet}()$ represents the last convolutional layer preceding the fully connected layers in our RaceNet model, which has been trained on the FairFace dataset.

Finally, the objective function of the training process is given by~\ref{eq:L-entire}:
\begin{align}
\mathcal{L}_{\text{RA-GAN}}(x, x^\prime, F_{mix}) = 
\lambda_{l2}\mathcal{L}_{l2}(x,x^\prime) + \lambda_{id}\mathcal{L}_{id}(x,x^\prime) + \notag\\ 
\lambda_{w_{norm}}\mathcal{L}_{w_{norm}}(F_{mix}) + \lambda_{Aging}\mathcal{L}_{Aging}(x, x^\prime) + \notag \\
\lambda_{race}\mathcal{L}_{race}(x, x^\prime))
\label{eq:L-entire}
\end{align}
Finally, the pseudo-code of our proposed RA-GAN, written in Algorithms~\ref{alg:ragan} and~\ref{alg:ragan-forward}, helps clarify the connection between various parts and the execution process.
\begin{algorithm}
\caption{\textbf{The Proposed RA-GAN}}
\label{alg:ragan}
\begin{algorithmic}[1]

\ForAll{$(x, \alpha_s, \alpha_t) \in \text{dataloader}$}

\State $\text{Move all data to GPU}$
\State $x^{\prime}, F_{mix} \gets \text{Forward}(x, \alpha_s, \alpha_t)$

\Statex
\State $\begin{aligned}[t]
	\mathcal{L}_{\text{RA-GAN}}(x, x^\prime, F_{\text{mix}}) = 
	&\ \lambda_{l2} \mathcal{L}_{l2}(x,x^\prime) 
	+ \lambda_{id} \mathcal{L}_{id}(x,x^\prime) \\
	&+ \lambda_{w_{\text{norm}}} \mathcal{L}_{w_{\text{norm}}}(F_{\text{mix}}) \\
	&+ \lambda_{\text{Aging}} \mathcal{L}_{\text{Aging}}(x, x^\prime) \\
	&+ \lambda_{race}\mathcal{L}_{\text{race}}(x, x^\prime)
\end{aligned}$
\Statex
\Statex
\State $x^r, F_{mix} \gets \text{Forward}(x^\prime, \alpha_t, \alpha_s)$

\Statex \Comment{$x^r$ is the reconstructed image}
\State $\begin{aligned}[t]
	\mathcal{L}^{r}_{\text{RA-GAN}}(x^r, x, F_{\text{mix}}) = 
	&\ \lambda_{l2} \mathcal{L}_{l2}(x^r,x) 
	+ \lambda_{id} \mathcal{L}_{id}(x^r,x) \\
	&+ \lambda_{w_{\text{norm}}} \mathcal{L}_{w_{\text{norm}}}(F_{\text{mix}}) \\
	&+ \lambda_{\text{Aging}} \mathcal{L}_{\text{Aging}}(x^r, x) \\
	&+ \lambda_{race}\mathcal{L}_{\text{race}}(x^r, x)
\end{aligned}$
\Statex

\State $\text{set gradients of parameters in optimizers to zero.}$
\State $\mathcal{L}_{RA-GAN}.backward()$

\State $\mathcal{L}^{r}_{RA-GAN}.backward()$
\State $\text{optimizer.step()}$

\EndFor
\end{algorithmic}
\end{algorithm}

\begin{algorithm}
\caption{\textbf{The Forward Method}}
\label{alg:ragan-forward}
\begin{algorithmic}[2]
\Require
\Statex $\triangleright$~$x$ is a normalized facial image
\Statex $\triangleright$ $\alpha_s$ is the age of $x$ (source age)
\Statex $\triangleright$ $\alpha_t$ is the target age of $x$

\Ensure
\Statex $x^{\prime}$
\Statex $F_{mix}$
\Statex

\State $x_{age} \gets x\parallel \alpha_{t}$ 
\State $S_{age} \gets E_{age}(x_{age})$ 
\State $S_{face} \gets E_{race}(x)$ 
\State $F_{mix} \gets F(stack(S_{age},S_{face}))$ 
\State $x^{\prime} \gets \text{StyleGAN-V2}(F_{mix})$

\State \Return $x^{\prime}, F_{mix}$
\end{algorithmic}
\end{algorithm}

\subsection{Kinship Verification with Aged Faces}
\label{subsec:kinship-ragan}
Our idea is to modify the facial ages of individuals using generative models and then verify their family connections to mitigate the age-varying challenge in KV. There are some obstacles to overcome. The faces in the KV benchmark datasets are cropped, and their heads and necks are missing from their images. With this restriction, facial images cannot be given to GANs, as they have been trained on full-face photos. If facial images were not full, corrupted faces would be received.

\begin{figure*}[h]
\begin{center}
\includegraphics[width=\textwidth]{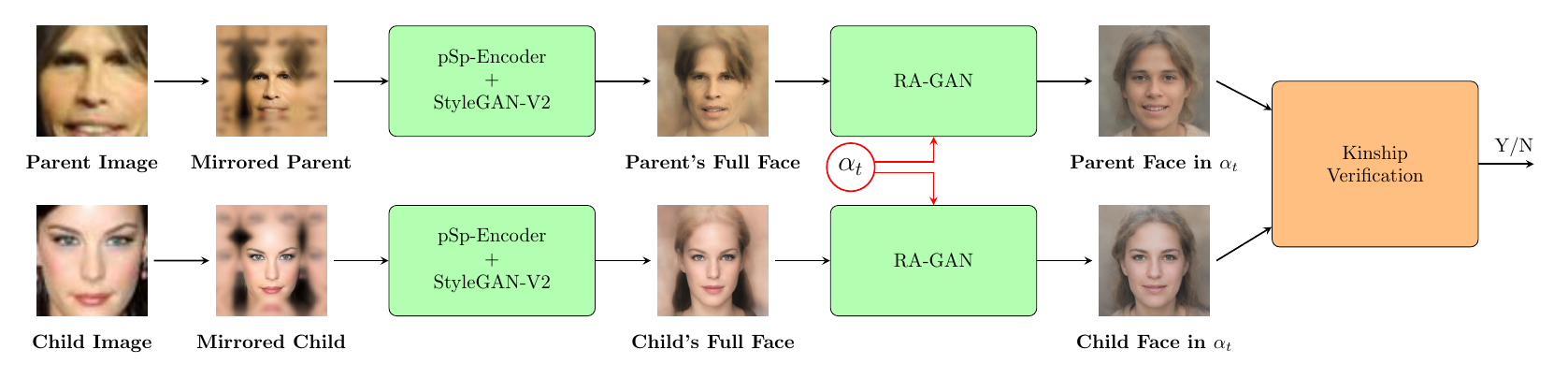}
\caption{The general diagram of kinship verification with aged facial images. The $\alpha$ parameter is the desired age at which we want to transform facial images.}
\label{fig:kinship-aged}
\end{center}
\end{figure*}

First, we attempt to convert cropped images to full-face images using the autoencoder (pSp-Encoder and StyleGAN-V2). Before feeding those images to the autoencoder, we apply mirror augmentation to each image. Second, these full-face images and the desired age are given to our RA-GAN. Finally, the converted images can be given to the state-of-the-art KV model to verify their kinship. All these steps are formulated in~(\ref{eq:kinship-fullface}) and illustrated in Fig.~\ref{fig:kinship-aged}.

\begin{align}
x_{fullface}^p &= \text{Autoencoder}\left( \text{Mirror}(x^p) \right) \notag \\
x_{fullface}^c &= \text{Autoencoder}\left( \text{Mirror}(x^c) \right) \notag \\
x'^p  &= \text{RA-GAN} \left( x_{fullface}^p ,\ {\alpha}_{t} \right) \notag \\
x'^c  &= \text{RA-GAN} \left( x_{fullface}^c ,\ {\alpha}_{t} \right) \notag \\
decision  &= \text{Kinship-Verification} \left( x'^p , x'^c \right) \notag \\
\label{eq:kinship-fullface}
\end{align}
The $\alpha$ parameter is the desired age at which we want to transform facial images.

\section{Face Aging Experiment}
\label{sec:experiment-face-aging}
In this section, we present the results of our extensive experimentation and comparison of our proposed RA-GAN with state-of-the-art algorithms, exploring its effectiveness in age transformation, latent space manipulation, and preservation of race and identity. All experiments were performed on a system with an RTX 4090 GPU.  

\subsection{Face Aging Dataset and Pre-processing}
\begin{figure*}[htbp]
\centering
\begin{minipage}[t]{0.3\textwidth}
\vspace{0pt}
\centering
\includegraphics[width=\linewidth]{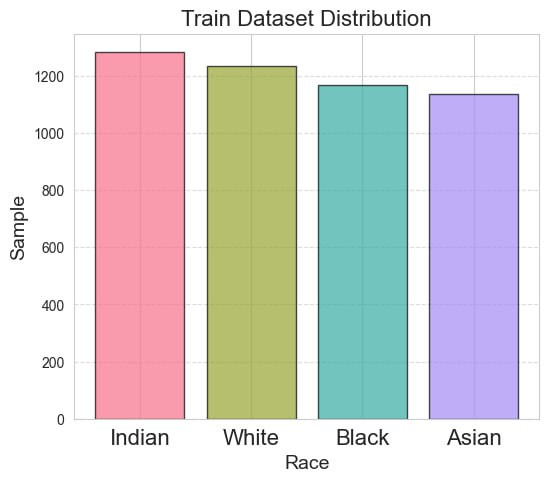}
\end{minipage}
\hspace{0.4cm}
\begin{minipage}[t]{0.3\textwidth}
\vspace{0pt}
\centering
\includegraphics[width=\linewidth]{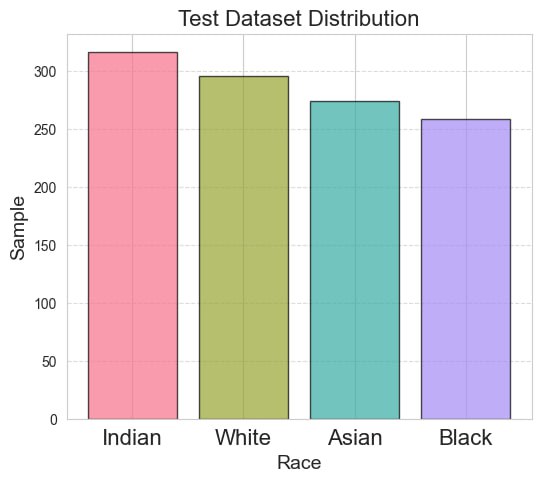}
\end{minipage}
\hspace{0.4cm}
\begin{minipage}[t]{0.3\textwidth}
\vspace{0.8cm}
\centering
\resizebox{\linewidth}{!}{%
	\begin{tabular}{lcc}
		\toprule
		Race   & Train   & Test \\ 
		\midrule
		Indian & 1284                       & 316  \\
		White  & 1235                       & 295  \\
		Asian  & 1138                       & 274  \\
		Black  & 1170                       & 258  \\ 
		\bottomrule
	\end{tabular}
}
\end{minipage}
\caption{The dataset is constructed from the UTKFace. The number of various race samples for training and test sets is mentioned.}
\label{fig:dataset_race_map}
\end{figure*}

\begin{figure*}[h]
\begin{center}
\includegraphics[width=0.35\textwidth]{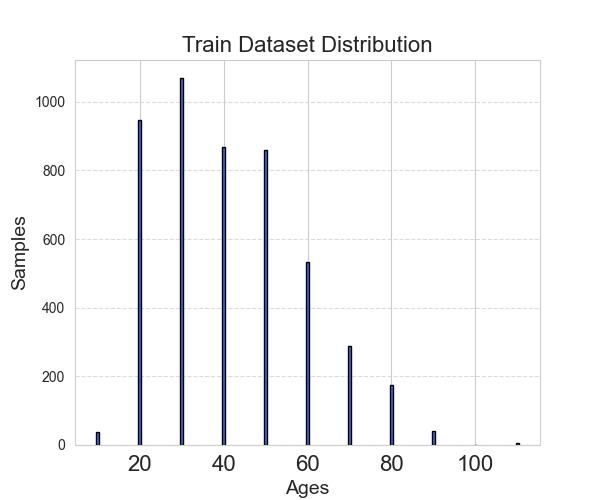}
\includegraphics[width=0.35\textwidth]{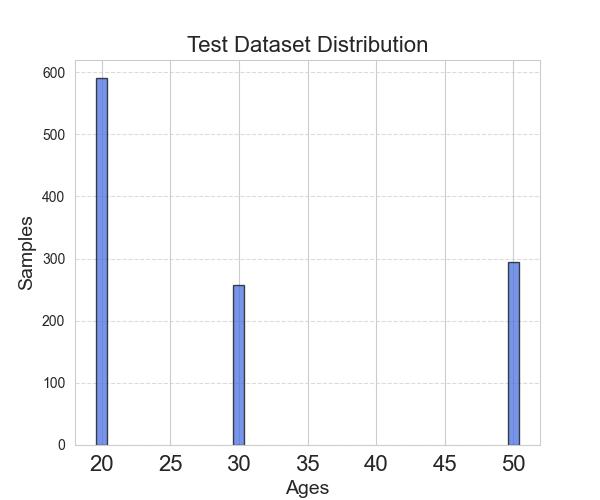}
\caption{The age distributions of the constructed datasets from the UTKFace are illustrated. }
\label{fig:dataset_age_map}
\end{center}
\end{figure*}

To train our model to generalize well across various races and poses and find the optimal manifold, we developed a dataset from the UTKFace dataset~\cite{zhifei2017cvpr}, which includes a wide variety of images, depicted in Fig.~\ref{fig:dataset_race_map}, representing four races: Indian, White, Black, and Asian. These collected datasets are uniform in terms of race, but not in terms of age, as illustrated in Fig.~\ref{fig:dataset_age_map}. If we want the datasets to be uniform based on both race and age, the number of images will be reduced. 

Furthermore, we denoised the images using GFPGAN~\cite{wang2021towards} and obtained the photos with a resolution of $1024\times1024$.
Since GFPGAN and StyleGAN-V2 were trained on the FFHQ dataset, the distribution of latent codes in our denoised dataset will become similar to that of StyleGAN-V2.

\subsection{Comparison with Face Aging Methods}
\begin{figure*}[tp]
\begin{center}
\begin{tabular}{>{\bfseries}cc} 
	~ & 
		\begin{tabular}{@{}c@{\hspace{1.3cm}}c@{\hspace{1.3cm}}c@{\hspace{1.3cm}}c@{\hspace{1.3cm}}c@{\hspace{1.3cm}}c@{\hspace{1.3cm}}c@{\hspace{0.8cm}}c@{}}
			20 & 30 & 40 & 50 & 60 & 70 & 80 & Source
		\end{tabular}
		\\
		\rotatebox{90}{\textbf{~CUSP~~~SAM~~~RA-GAN}} &
		\includegraphics[width=\textwidth]{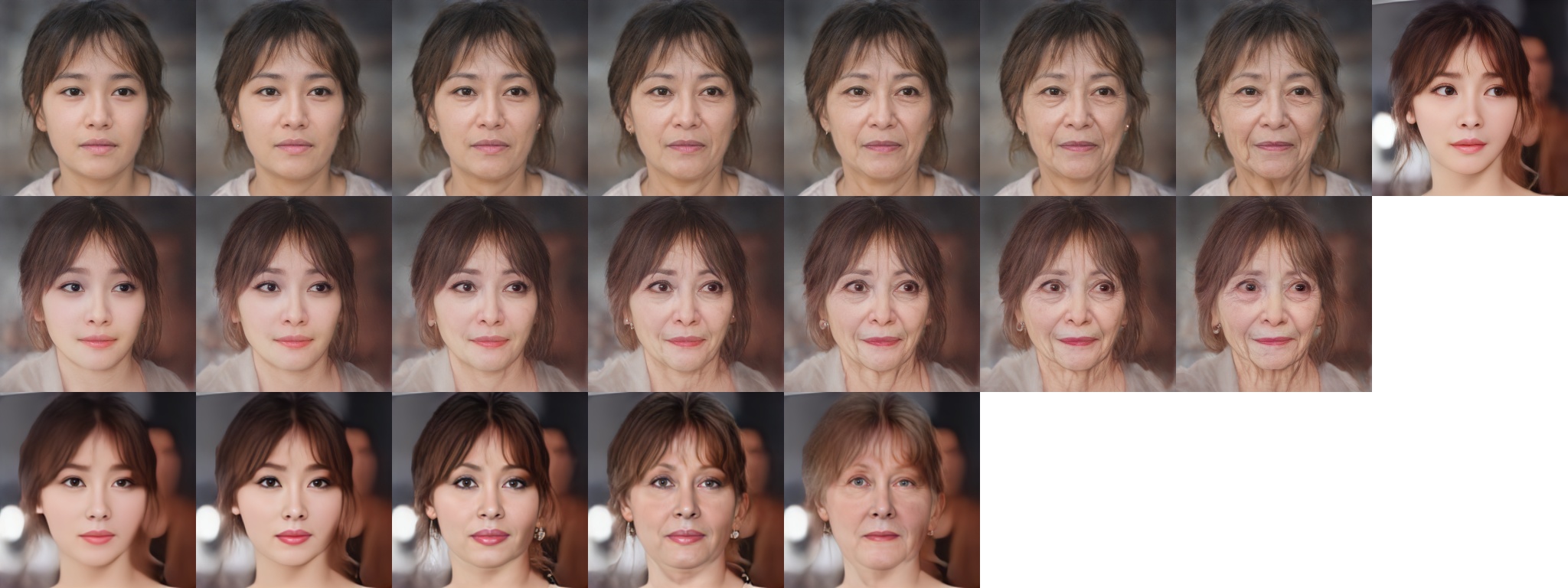}
		\\
	\end{tabular}
	\caption{Outputs of RA-GAN, SAM-GAN, and CUSP-GAN in order from top to bottom. The right upper picture is the source picture, and the rest were produced by models from the 20 to 80 age group.}
	\label{fig:outputs-ragan-sam-cusp}
\end{center}
\end{figure*}

We evaluate our RA-GAN against SAM-GAN~\cite{alaluf2021only} and CUSP~\cite{gomez2022custom}, using their official implementations.
We cover an age range of 20 to 80 years with a 10-year step, except for CUSP, covering the age group until 60. 
Face aging models transform facial images to different ages; Still, the identities and racial characteristics of individuals may change in some cases, as shown in one identity in Fig.~\ref{fig:outputs-ragan-sam-cusp}. This incorrect transformation of facial images can affect the accuracy of the kinship determination. To assess their ability to preserve the race and identity, specific scores will be utilized. 

\subsubsection{Generating Realistic Images}
\begin{figure*}[htp]
\centering
\resizebox{.9\linewidth}{!}{
	\begin{tabular}{>{\bfseries}c | c} 
		M & 
			\begin{tabular}{c@{\hspace{0.5cm}}c@{\hspace{1.1cm}}c@{\hspace{1.3cm}}c@{\hspace{1.3cm}}c@{\hspace{1.3cm}}c@{\hspace{1.3cm}}c@{\hspace{0.8cm}}c@{}}
				
				20 & 30 & 40 & 50 & 60 & 70 & 80 & Source
			\end{tabular}
			\\
			
			\rotatebox{90}{\textbf{~CUSP~~~SAM~~~RA-GAN}} &
			\includegraphics[width=0.9\textwidth]{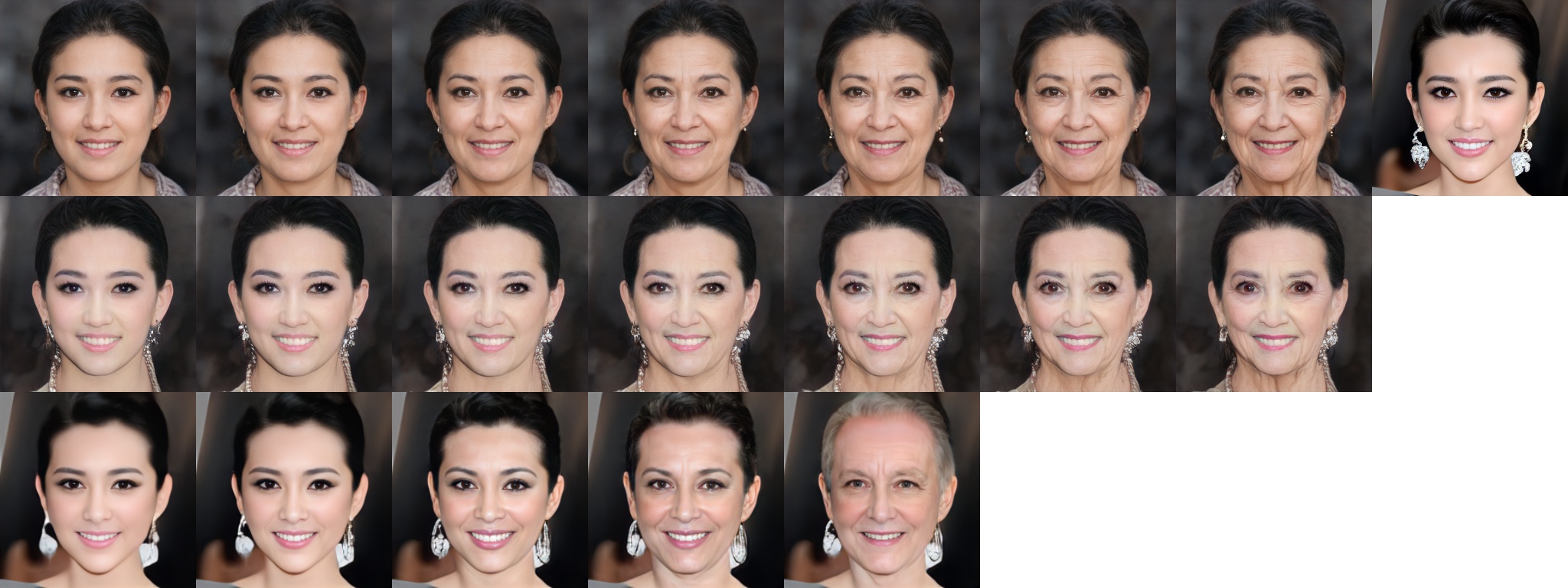} \\
			
			\rotatebox{90}{\textbf{~CUSP~~~SAM~~~RA-GAN}} &
			\includegraphics[width=0.9\textwidth]{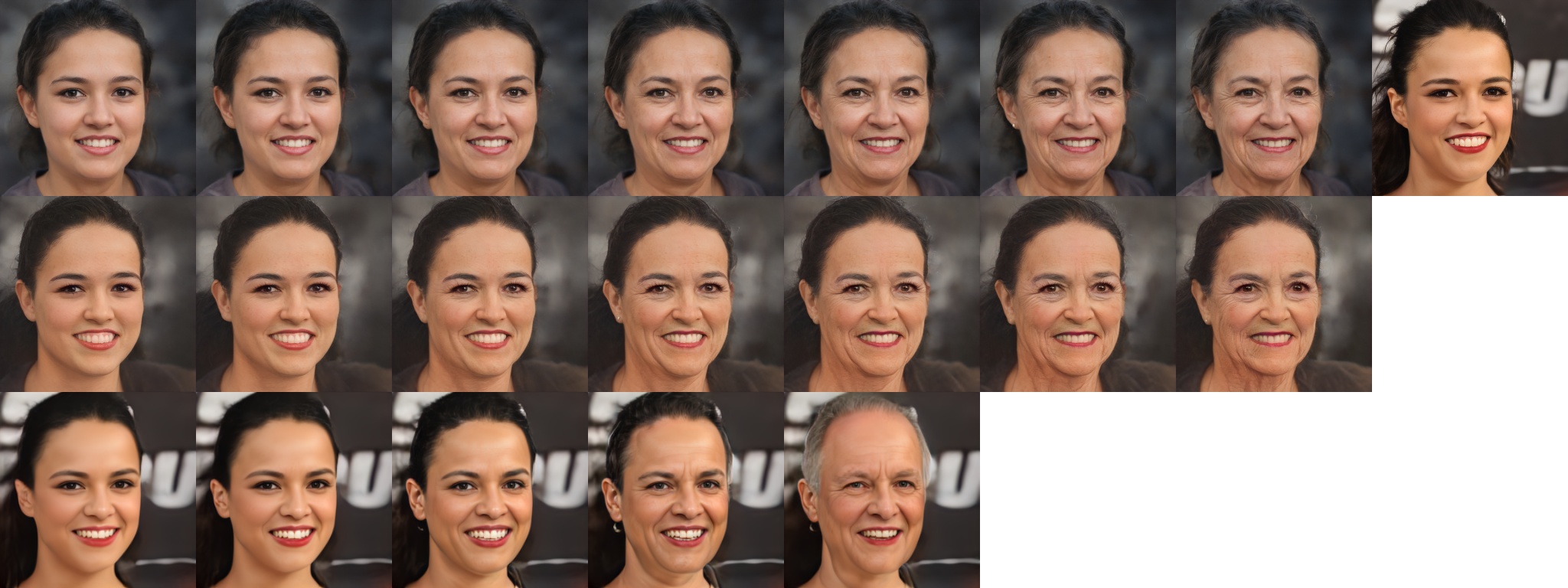} \\
			
			\rotatebox{90}{\textbf{~CUSP~~~SAM~~~RA-GAN}} &
			\includegraphics[width=0.9\textwidth]{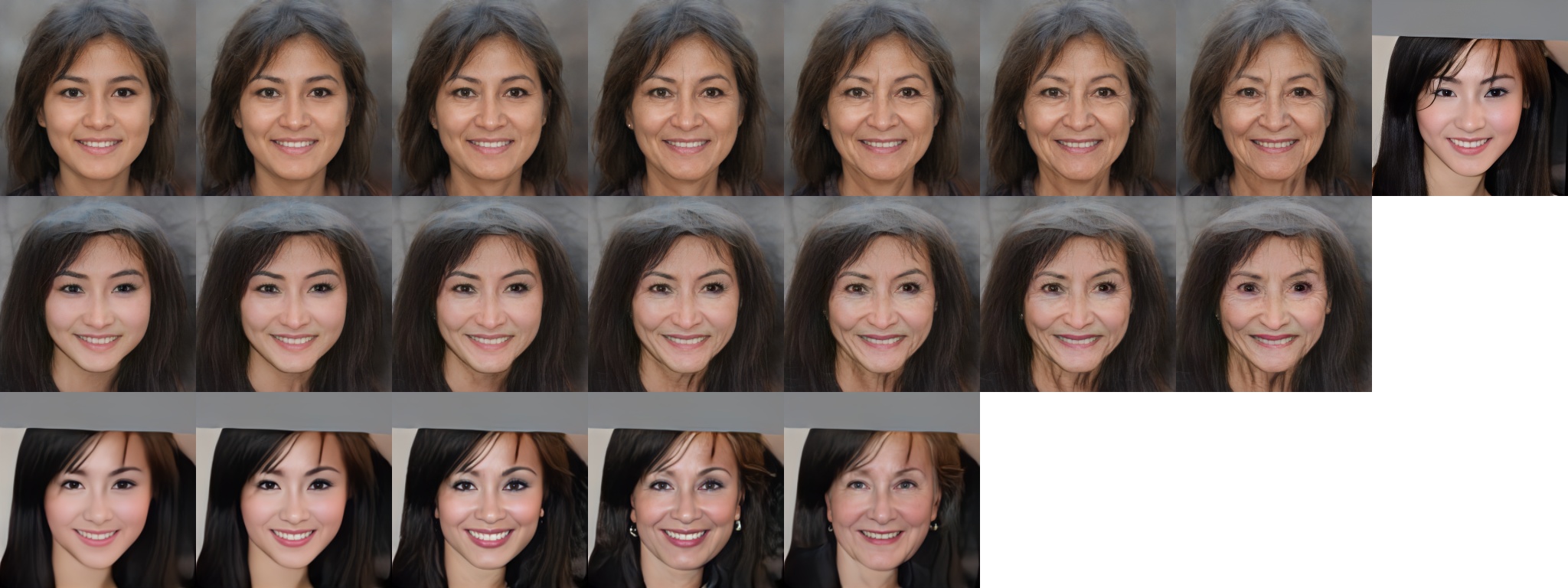} \\
			
			\rotatebox{90}{\textbf{~CUSP~~~SAM~~~RA-GAN}} &
			\includegraphics[width=0.9\textwidth]{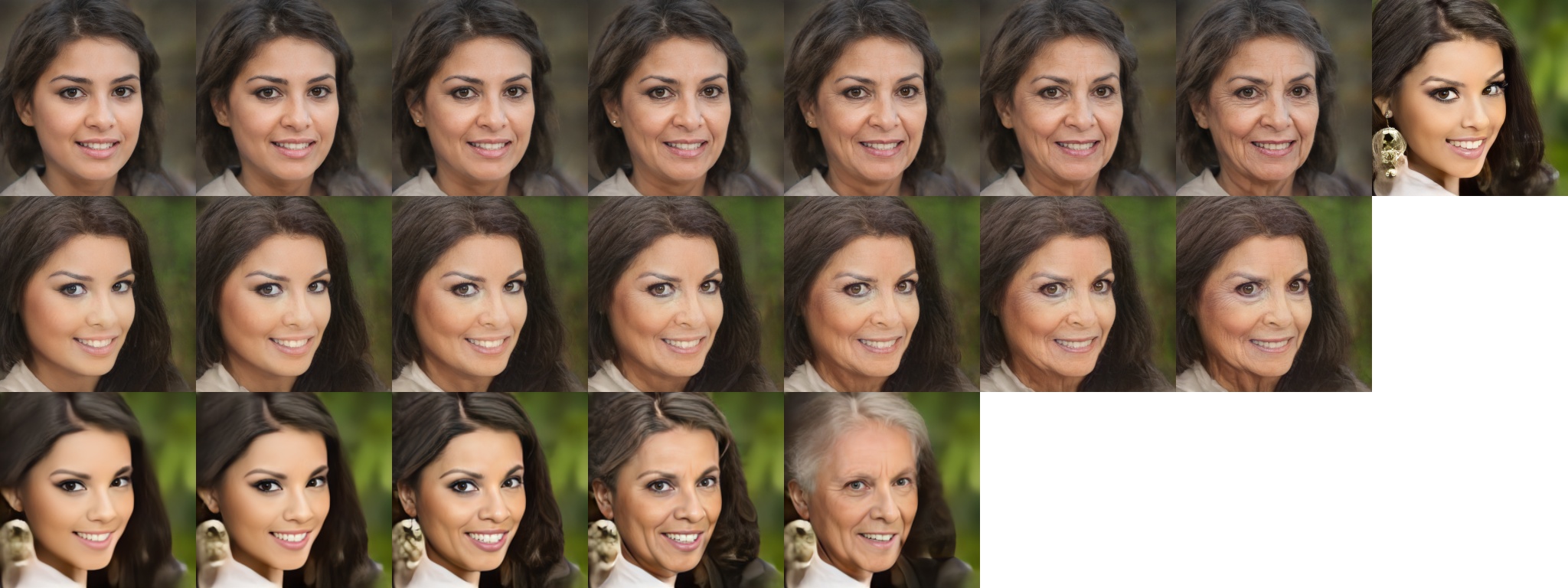} \\
		\end{tabular}
	}
	\caption{Comparison of synthesized images of RA-GAN, SAM-GAN, and CUSP with left-aligned age labels.}
	\label{fig:race_identity_sample_1}
\end{figure*}

Generative models can be compared with each other based on their capability to produce visually realistic images, as well as by specific measures such as racial preservation and correct age transformation.
We provide a visual comparison between SAM-GAN, CUSP, and RA-GAN in terms of geometrical face shape, race, and texture, using some pictures from models in Fig.~\ref {fig:race_identity_sample_1}.
The geometrical face shape across various ages can be captured well in models. 
The race of images generated by SAM-GAN and CUSP is changed but maintained by RA-GAN. 
In addition, some facial components, such as the jaw and forehead, have been modified by SAM-GAN; for example, the shape of the jaw is not aligned with all other facial components in some cases. Furthermore, CUSP changes race drastically in older ages, especially at 60.
In contrast, our RA-GAN can preserve the relationship between all components of the face, such as the eyes, forehead, chin, and jaw. 
Furthermore, skin texture is modeled more naturally in RA-GAN than in SAM-GAN since RA-GAN can produce better old facial images. 

\subsubsection{Race and Identity Preservation}
Our aim is to preserve racial characteristics and identity attributes throughout the aging transformation process. 
It requires precisely mapping the model's output to the input and achieving an acceptable generalization despite unfavorable conditions, such as various poses and aging effects. These attributes can help face applications, such as KV to achieve higher results.

We compare our RA-GAN, CUSP, and SAM-GAN visually by illustrating pictures of some cases in Fig.~\ref{fig:race_identity_sample_1}, similar to \cite{alaluf2021only}. SAM-GAN cannot preserve race because of changes in eyes, chins, and eyebrows at various ages. CUSP also cannot preserve race, especially at older ages, whereas RA-GAN can maintain it by considering race features through its RaceNet.

\subsection{Hyper-parameters and Parameters}
Most hyperparameters and parameters of our RA-GAN are similar to those of SAM-GAN. Parameters of the learning rate, Betas, and weight decay for the first optimizer are 1e-7, (0.9,0.99), and 0; for the reconstructed phase, they are 5e-5, (0.5, 0.99), and 1e-5, respectively. We run with a batch size of four images with a resolution of (256,256). The $\lambda$ hyper-parameters $\lambda_{id}$, $\lambda_{l2}$, $\lambda_{w_{norm}}$, $\lambda_{Aging}$, and $\lambda_{race}$ are 0.1, 0.25, 0.005, 5, and 3, respectively.

\subsection{Race Preservation in Detail}
This section demonstrates quantitatively the ability of models to preserve the racial features of aged images using metrics such as their confusion matrix, accuracy, recall, and F-score. 
First, we train ResNet-34 on the Fairface dataset introduced in~\cite{karkkainen2021fairface}, which comprises four racial classes: White, Black, Asian, and Indian.
Then, we transform the age of test images using SAM-GAN, CUSP, and RA-GAN and measure their race using our pre-trained ResNet-34.
Additionally, the dlib package is utilized to crop the facial region of the synthesized images; the same area is fed into the pre-trained ResNet-34 on the Fairface dataset.
The resulting confusion matrices are illustrated in Fig.~\ref{fig:race_confusion_matrices-1} and Fig.~\ref{fig:race_confusion_matrices-2}. 

To clarify the effectiveness of our method, we calculate the racial accuracy of models for each age group from confusion matrices. We present the results in Table~\ref{table:accuracy-race}, which show that RA-GAN outperforms SAM-GAN in various age groups, achieving higher racial accuracy and an average improvement of 13.14 across all age groups compared to SAM-GAN. Furthermore, RA-GAN noticeably outperforms CUSP in the older age groups and is nearly equal in the younger age groups.

In addition, we have mentioned other measures, such as precision, recall, and F-score, for RA-GAN, SAM-GAN, and CUSP-GAN in Tables~\ref{table:ra-gan-fscore}, \ref{table:sam-gan-fscore}, and \ref{table:cups-gan-fscore}, respectively. We compare the F-scores of the RA-GAN, SAM-GAN, and CUSP-GAN models across all ages and portray the improvement amount for each race in Table~\ref{table:fscore-race-comparison-ra-to-sam}. 
RA-GAN outperforms SAM-GAN in preserving racial diversity across all age groups and can improve older age groups, specifically in comparison to CUSP-GAN.

\subsection{Identity Preservation Score}
Identity preservation in GANs is essential for several reasons. First, transformed facial images must be identical to the original; otherwise, drastic changes can render a facial image unrecognizable, making it appear to be that of another person. Furthermore, in the following processing task, here, KV, the identity of individuals must be the same as the original so that kinship verification can be reliable. To measure the amount of identity preservation by models, we use the cosine similarity between the ArcFace CNN features of the original image and those of its synthesized counterparts.
As shown in Table~\ref{table:error-identity-perservation}, the cosine similarity of the models denotes that  RA-GAN can preserve the identity similar to SAM-GAN. However, CUSP-GAN performs better for the younger age group, whereas our RA-GAN outperforms it for the older age group.

\begin{table*}[t]
	\centering
	\caption{The racial accuracy of RA-GAN and SAM-GAN}
	\label{table:accuracy-race}
	\begin{adjustbox}{width=0.85\textwidth}
		\begin{tabular}{|cccccc|}
			\toprule
			Age & RA-GAN(\%) & SAM-GAN(\%) & CUSP-GAN(\%) & 
			\begin{tabular}{@{}c@{}}Improvement \\  RA to SAM\end{tabular} &
			\begin{tabular}{@{}c@{}}Improvement \\  RA to CUSP\end{tabular} \\
			\midrule
			20 & 75.59 & 65.79 & 78.21 &9.8 & -2.62\\
			\midrule
			30 & 77.42 & 67.10 & 78.91 &10.32 & -1.49 \\
			\midrule
			40 & 76.64 & 63.34 & 78.82 &13.3  & -2.18\\
			\midrule
			50 & 71.39 & 55.38 & 70.42 &16.01 & 0.97 \\
			\midrule
			60 & 62.82 & 48.38 & 53.71 &14.44  &  9.1\\
			\midrule
			70 & 57.83 & 42.43 & - &15.4  &  -\\
			\midrule
			80 & 49.26 & 36.57 & - &12.69  & -\\
			\bottomrule
		\end{tabular}
	\end{adjustbox}
\end{table*}

\begin{table*}[htbp]
	\centering
	\caption{RA-GAN results}
	\label{table:ra-gan-fscore}
	\begin{adjustbox}{width=\textwidth}
		\begin{tabular}{ cc }
			\toprule
			\begin{tabular}{|c||c|c|c|}
				\hline
				\multicolumn{4}{|c|}{\textbf{Indian}} \\
				\hline
				\textbf{Ages} & \textbf{Precision} & \textbf{Recall} &
				\textbf{F1-score} \\ 
				\hline
				20 & 0.73 & 0.78 & 0.76 \\ 
				30 & 0.77 & 0.79 & 0.78 \\ 
				40 & 0.79 & 0.72 & 0.75 \\ 
				50 & 0.80 & 0.63 & 0.70 \\ 
				60 & 0.74 & 0.48 & 0.58\\ 
				70 & 0.66 & 0.38 & 0.48 \\ 
				80 & 0.57 & 0.22 & 0.31\\
				\hline
				\hline
				\multicolumn{4}{|c|}{\textbf{Asian}} \\
				\hline
				\textbf{Ages} & \textbf{Precision} & \textbf{Recall} &
				\textbf{F1-score} \\ 
				\hline
				20 & 0.82 & 0.77 & 0.80 \\ 
				30 & 0.82 & 0.76  & 0.79 \\ 
				40 & 0.84 & 0.73 & 0.78 \\ 
				50 & 0.82 & 0.62 & 0.71 \\ 
				60 & 0.75 & 0.54 & 0.63\\ 
				70 & 0.71 & 0.53 & 0.61 \\ 
				80 & 0.54 & 0.57 & 0.55\\ 
				\hline
			\end{tabular} &
			\begin{tabular}{|c||c|c|c|}
				\hline
				\multicolumn{4}{|c|}{\textbf{Black}} \\
				\hline
				\textbf{Ages} & \textbf{Precision} & \textbf{Recall} &
				\textbf{F1-score} \\ 
				\hline
				20 & 0.90 & 0.59 & 0.71 \\ 
				30 & 0.92 & 0.68 & 0.78 \\ 
				40 & 0.92 & 0.71 & 0.80 \\ 
				50 & 0.93 & 0.67 & 0.78 \\ 
				60 & 0.94 & 0.57 & 0.71\\ 
				70 & 0.94 & 0.46 & 0.62 \\ 
				80 & $\sim$1.00 & 0.25 & 0.40\\ 
				\hline
				\hline
				\multicolumn{4}{|c|}{\textbf{White}} \\
				\hline
				\textbf{Ages} & \textbf{Precision} & \textbf{recall} &
				\textbf{F1-score} \\ 
				\hline
				20 & 0.67 & 0.86 & 0.75 \\ 
				30 & 0.68 & 0.86 & 0.76 \\ 
				40 & 0.63 & 0.89 & 0.74 \\ 
				50 & 0.55 & 0.94 & 0.69 \\ 
				60 & 0.47 & 0.93 & 0.62\\ 
				70 & 0.44 & 0.93 & 0.60 \\ 
				80 & 0.41 & 0.93 & 0.57\\ 
				\hline
			\end{tabular}
		\end{tabular}
	\end{adjustbox}
\end{table*}


\begin{table*}[]
	\centering
	\caption{SAM-GAN reports}
	\label{table:sam-gan-fscore}
	\begin{adjustbox}{width=\textwidth}
		\begin{tabular}{ cc }
			\toprule
			\begin{tabular}{|c||c|c|c|}
				\hline
				\multicolumn{4}{|c|}{\textbf{Indian}} \\
				\hline
				\textbf{Ages} & \textbf{Precision} & \textbf{Recall} &
				\textbf{F1-score} \\ 
				\hline
				20 & 0.74 & 0.40 & 0.52 \\ 
				30 & 0.78 & 0.41 & 0.53 \\ 
				40 & 0.80 & 0.38 & 0.52 \\ 
				50 & 0.74 & 0.28 & 0.41 \\ 
				60 & 0.65 & 0.24 & 0.35\\ 
				70 & 0.51 & 0.16 & 0.25 \\ 
				80 & 0.40 & 0.09 & 0.15\\ 
				\hline
				\hline
				\multicolumn{4}{|c|}{\textbf{Asian}} \\
				\hline
				\textbf{Ages} & \textbf{Precision} & \textbf{Recall} &
				\textbf{F1-score} \\ 
				\hline
				20 & 0.76 & 0.82 & 0.79 \\ 
				30 & 0.84 & 0.78 & 0.81 \\ 
				40 & 0.87 & 0.67 & 0.76 \\ 
				50 & 0.87 & 0.50 & 0.63 \\ 
				60 & 0.84 & 0.38 & 0.52\\ 
				70 & 0.77 & 0.31 & 0.44 \\ 
				80 & 0.63 & 0.29 & 0.40\\ 
				\hline
			\end{tabular} &
			\begin{tabular}{|c||c|c|c|}
				\hline
				\multicolumn{4}{|c|}{\textbf{Black}} \\
				\hline
				\textbf{Ages} & \textbf{Precision} & \textbf{Recall} &
				\textbf{F1-score} \\ 
				\hline
				20 & 0.97 & 0.48 & 0.64 \\ 
				30 & 0.97 & 0.55 & 0.70 \\ 
				40 & 0.98 & 0.51 & 0.67 \\ 
				50 & 0.98 & 0.45 & 0.61 \\ 
				60 & 0.98 & 0.31 & 0.48\\ 
				70 & 0.98 & 0.22 & 0.36 \\ 
				80 & 0.95 & 0.07 & 0.13\\
				\hline
				\hline
				\multicolumn{4}{|c|}{\textbf{White}} \\
				\hline
				\textbf{Ages} & \textbf{Precision} & \textbf{Recall} &
				\textbf{F1-score} \\ 
				\hline
				20 & 0.51 & 0.94 & 0.66 \\ 
				30 & 0.49 & 0.97 & 0.65 \\ 
				40 & 0.45 & 0.98 & 0.61 \\ 
				50 & 0.39 & 0.99 & 0.56 \\ 
				60 & 0.36 & 0.99 & 0.53\\ 
				70 & 0.33 & 0.99 & 0.50 \\ 
				80 & 0.32 & 0.99 & 0.48\\ 
				\hline
			\end{tabular}
		\end{tabular}
	\end{adjustbox}
\end{table*}


\begin{table*}[]
	\centering
	\caption{CUSP-GAN reports}
	\label{table:cups-gan-fscore}    
	\begin{adjustbox}{width=\textwidth}
		\begin{tabular}{ cc }
			\toprule
			\begin{tabular}{|c||c|c|c|}
				\hline
				\multicolumn{4}{|c|}{\textbf{Indian}} \\
				\hline
				\textbf{Ages} & \textbf{Precision} & \textbf{Recall} &
				\textbf{F1-score} \\ 
				\hline
				20 & 0.80 & 0.73 & 0.77 \\ 
				30 & 0.78 & 0.78 & 0.78 \\ 
				40 & 0.73 & 0.87 & 0.79 \\ 
				50 & 0.77 & 0.81 & 0.79 \\ 
				60 & 0.82 & 0.32 & 0.46\\ 
				\hline
				\hline
				\multicolumn{4}{|c|}{\textbf{Asian}} \\
				\hline
				\textbf{Ages} & \textbf{Precision} & \textbf{Recall} &
				\textbf{F1-score} \\ 
				\hline
				20 & 0.83 & 0.86 & 0.85 \\ 
				30 & 0.72 & 0.95 & 0.82 \\ 
				40 & 0.92 & 0.67 & 0.78 \\ 
				50 & 0.88 & 0.28 & 0.42 \\ 
				60 & 0.63 & 0.30 & 0.40\\ 
				\hline
			\end{tabular} &
			\begin{tabular}{|c||c|c|c|}
				\hline
				\multicolumn{4}{|c|}{\textbf{Black}} \\
				\hline
				\textbf{Ages} & \textbf{Precision} & \textbf{Recall} &
				\textbf{F1-score} \\ 
				\hline
				20 & 0.83 & 0.82 & 0.82 \\ 
				30 & 0.92 & 0.76 & 0.83 \\ 
				40 & 0.81 & 0.86 & 0.84 \\ 
				50 & 0.80 & 0.87 & 0.83 \\ 
				60 & 0.94 & 0.62 & 0.74\\ 
				\hline
				\hline
				\multicolumn{4}{|c|}{\textbf{White}} \\
				\hline
				\textbf{Ages} & \textbf{Precision} & \textbf{Recall} &
				\textbf{F1-score} \\ 
				\hline
				20 & 0.68 & 0.72 & 0.70 \\ 
				30 & 0.80 & 0.67 & 0.73 \\ 
				40 & 0.75 & 0.74 & 0.75 \\ 
				50 & 0.56 & 0.84 & 0.67 \\ 
				60 & 0.38 & 0.92 & 0.54\\ 
				\hline
			\end{tabular}
		\end{tabular}
	\end{adjustbox}
\end{table*}


\begin{table*}[]	
	\centering
	\caption{This portrays the RA-GAN improvement of F-scores in comparison with SAM-GAN on average across all ages for various races.}
	\label{table:fscore-race-comparison-ra-to-sam}
	\begin{tabular}{cccc}
		\toprule
		Indian & Black & White & Asian  \\
		\midrule
		23.28   & 17.28 & 10.57  & 7.42\\
		\bottomrule
	\end{tabular}
\end{table*}

\begin{table*}[ht]
	\centering
	\caption{The Identity Preservation Scores of RA-GAN, SAM-GAN, and CUSP.}
	\label{table:error-identity-perservation}
	\begin{adjustbox}{width=0.65\textwidth}
		\begin{tabular}{lcccccc}
			\toprule
			\multirow{2}{*}{Age} 
			& \multicolumn{2}{c}{RA-GAN} 
			& \multicolumn{2}{c}{SAM-GAN}  
			& \multicolumn{2}{c}{CUSP-GAN} \\
			\cmidrule(lr){2-3} \cmidrule(lr){4-5} \cmidrule(lr){6-7}
			& Mean & Std & Mean & Std & Mean & Std \\
			\midrule
			20 & 0.494 & 0.089 & 0.607 & 0.083 & 0.673 & 0.088 \\
			30 & 0.488 & 0.095 & 0.559 & 0.091 & 0.701 & 0.078 \\
			40 & 0.463 & 0.099 & 0.493 & 0.091 & 0.628 & 0.121 \\
			50 & 0.427 & 0.102 & 0.454 & 0.096 & 0.460 & 0.143 \\
			60 & 0.413 & 0.099 & 0.443 & 0.096 & 0.353 & 0.144 \\
			70 & 0.414 & 0.095 & 0.440 & 0.094 & ---   & ---   \\
			80 & 0.404 & 0.094 & 0.419 & 0.095 & ---   & ---   \\
			\bottomrule
		\end{tabular}
	\end{adjustbox}
\end{table*}

\begin{table*}[ht]
	\centering
	\caption{The age MAE of RA-GAN, SAM-GAN, and CUSP.}
	\label{table:error-mae-age-mapping}
	\begin{adjustbox}{width=0.65\textwidth}
		\begin{tabular}{lcccccc}
			\toprule
			\multirow{2}{*}{Age} 
			& \multicolumn{2}{c}{RA-GAN} 
			& \multicolumn{2}{c}{SAM-GAN} 
			& \multicolumn{2}{c}{CUSP} \\
			\cmidrule(lr){2-3} \cmidrule(lr){4-5} \cmidrule(lr){6-7}
			& Mean & Std & Mean & Std & Mean & Std \\
			\midrule
			20 & 6.52 & 4.54 & 6.74 & 5.12 & 3.71 & 3.21 \\
			30 & 4.65 & 3.75 & 4.74 & 3.92 & 4.71 & 2.96 \\
			40 & 4.83 & 3.62 & 5.56 & 3.95 & 6.41 & 4.73 \\
			50 & 4.72 & 3.61 & 5.49 & 4.47 & 6.52 & 4.54 \\
			60 & 4.66 & 3.44 & 6.31 & 4.78 & 7.24 & 5.38 \\
			70 & 6.29 & 4.45 & 8.57 & 5.72 & ---  & ---  \\
			80 & 9.84 & 4.32 & 10.05 & 5.58 & ---  & ---  \\
			\bottomrule
		\end{tabular}
	\end{adjustbox}
\end{table*}

\subsection{Age Mean Absolute Error (MAE)}
We measure the age mapping using the MAE metric to determine whether they have been correctly transformed to the target age. To estimate the age of the images, we use the FACE++ API\footnote{Face++ API: https://www.faceplusplus.com/} because, according to CUSP experiments, the DEX pre-trained classifier~\cite{rothe2015dex} exhibits a greater bias toward younger age groups compared to FACE++. We randomly sample 1100 images for each age group from the UTKFACE dataset and then estimate the age of the synthesized images using FACE++. The difference between the target age and the estimation is calculated using the MAE.

RA-GAN's performance in age transformation is truly superior, consistently achieving lower error rates across all age groups compared to SAM-GAN, as shown in Table.~\ref{table:error-mae-age-mapping}.
Except for the 20-year age group, our RA-GAN can transform to the target age with less error than CUSP~\cite{gomez2022custom} for all other age groups, as shown in Table.~\ref{table:error-mae-age-mapping}.

\section{Kinship Experiment}
\label{sec:experiment}
We specify the kinship datasets used for age transformation in this section and then investigate the KV accuracy for each age group to prove that at a specific age, KV can be more effective and has higher accuracy than at other ages. Additionally, the age transformation of both parent and child images at the same age yields higher accuracy compared to when they are at different ages.

\subsection{Kinship Datasets}
The KinFaceW-I and KinFaceW-II datasets~\cite{lu2013neighborhood} are benchmarks in Kinship verification to compare different KV algorithms, consisting of four kinship relationships: Father-Son (FS), Father-Daughter (FD), Mother-Son (MS), and Mother-Daughter (MD). The number of image pairs for each relationship is 156, 134, 116, and 127 in the KinFaceW-I dataset. The KinFaceW-II dataset comprises 250 image pairs for each relationship. Most of its images were gathered from the Internet and are of celebrities. The parent and child images are from different sources under unconstrained conditions in the KinFaceW-I dataset and from the same source in the KinFaceW-II dataset. 

\subsection{Full-face Generation}
\begin{figure}[htp]
	\centering
	\renewcommand{\arraystretch}{0.5} 
	\begin{tabular}{@{} c@{} c@{} c@{} c@{}} 
		
		\includegraphics[width=0.20\linewidth]{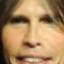}&
		\includegraphics[width=0.20\linewidth]{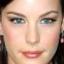}&
		\includegraphics[width=0.20\linewidth]{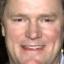}&
		\includegraphics[width=0.20\linewidth]{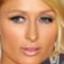} \\
		
		\includegraphics[width=0.20\linewidth]{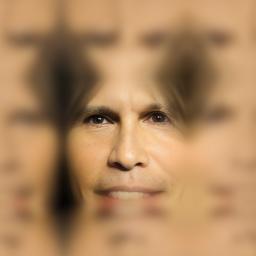}&
		\includegraphics[width=0.20\linewidth]{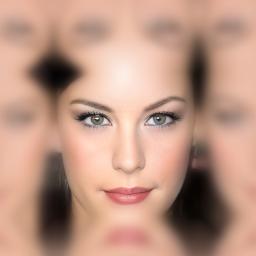}&
		\includegraphics[width=0.20\linewidth]{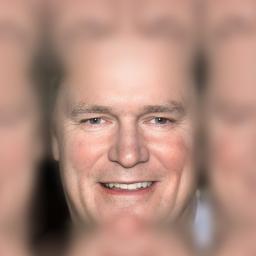}&
		\includegraphics[width=0.20\linewidth]{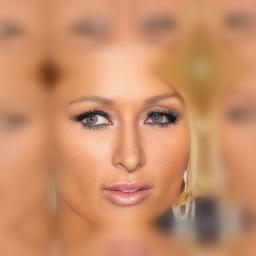} \\
	\end{tabular}
	\caption{Full-face generation of crop facial images. The top and bottom images are the original and mirrored ones, respectively.}
	\label{fig:full-face-generation}
\end{figure}

GANs accept full-face pictures, including all facial components, such as hair, chin, and ears, to produce proportional and authentic face images. Since KinFaceW-I images are cropped, and their full-face counterparts are not available, we attempt to convert them to full-face images by mirroring the cropped faces. In other words, cropped face images are padded by themselves and then given to the auto-encoder (pSp-encoder, StyleGAN-V2) to produce a full face. Some produced pictures are shown in Fig.~\ref{fig:full-face-generation}. Eventually, having full-face images, we can produce facial images at various ages using GANs like our RA-GAN.

\subsection{Kinship Verification Using Aged Synthesized Images}
Among the state-of-the-art algorithms of KV, the implementation code of D4ML~\cite{zhu2023distance} is available and utilized to evaluate the accuracy of pair images synthesized by RA-GAN. The training and test configurations follow those mentioned in the KinFaceW dataset.
All hyperparameters and parameters are similar to D4ML.
Results are shown in Table~\ref{table:kinship-results}, and the base result is based on full-face photos, which are not transformed. Other entries of results are based on facial images at that age.

We observe that the KV accuracy is generally increased in all four family relationships in the KinFaceW-I dataset when we transform facial images at the same age. The improvements in the FD, FS, MD, and MS kinships are 5.22$\%$, 5.15$\%$, 1.63$\%$, and 0.41$\%$, respectively. Furthermore, its improvements on the KinFaceW-II dataset in the FD, FS, and MS kinships are 2.9$\%$, 0.39$\%$, and 1.6$\%$, respectively.

\begin{table*}[t]
	\centering
	\caption{The KV results of the KinFaceW-I and II image-pairs synthesized by RA-GAN at various ages.}
	\label{table:kinship-results}
	\begin{tabular}{|c||llll||llll|}
		\toprule
		\multicolumn{1}{|c||}{\multirow{2}{*}{Ages}} & \multicolumn{4}{c||}{KinFaceW-I} & \multicolumn{4}{c|}{KinFaceW-II}  \\ 
		\multicolumn{1}{|c||}{}  & FD   & FS    & MD    &  {MS} &           FD   & FS     & MD   &  {MS} \\ \hline
		Base  & 72.1           & 74.04           & 83.86          & 71.52          &85.2           & 90             & 92.2  & 88.2 \\ \hline
		20    & 76.92          & 78.53           & \textbf{85.49} & 71.5           &87.35           & 90.19          & 90.39 &\textbf{89.8}  \\
		30    & 76.56          & 78.86           & 85.43          & 70.65          &87.73           & 89.8           & 90.19 &89.4 \\
		40    & \textbf{77.32} & 78.86           & 85.03          & 71.5           &87.35             & \textbf{90.39} & 90.8  &89.2\\
		50    & 76.56          & \textbf{79.19}  & 85.03          & \textbf{71.93} &87.73          & 89.8           & 91.4  &89.2 \\
		60    & 76.55          & 78.87           & 83.86          & 71.93          &\textbf{88.10}  & 89.4           & 91    &88  \\ \hline
		Improvement& 5.22          & 5.15            & 1.63           & 0.41       &2.9            & 0.39           & -     &1.6\\
		\bottomrule
	\end{tabular}
\end{table*}


\section{Conclusion and Future Work}
\label{sec:conclusion}
We aimed to examine the effect of same-age photos in kinship verification at various age groups and utilized face aging methods to overcome the unavailability of these photos. Since face aging methods often produce racially biased images, we propose an effective model to eliminate this bias as well as to preserve identity. In addition, we compiled a new dataset from the UTKFace dataset, which is balanced in terms of race.

Our proposed face aging model, RA-GAN, possesses the distinctive ability to synthesize racially unbiased facial photos through its new RACEpSp module and feature mixer module, surpassing SAM-GAN in all ages and CUSP-GAN in older ages in terms of racial accuracy and identity preservation. 
Our experiments on the KinFaceW-I dataset revealed an increase in the accuracy of kinship verification when parent-child images are converted to the same age. For instance, this conversion led to a substantial accuracy boost on the KinFaceW-I dataset for father-son and father-daughter, mother-son, and mother-daughter relations to 5.22, 5.12, 1.63, and 0.41, respectively, demonstrating the practical implications of our model. Furthermore, the results reveal that this conversion can enhance kinship accuracy when photos originate from various sources and were taken at different times, which is often the case in the real world, rather than having pictures under the same conditions all the time.

Moreover, we assert that the application of our method in fusing racial features in GANs has high potential to eradicate racial bias and generate images that closely resemble the original images' racial categories.

\section{Data availability}
The Datasets used in this paper are publicly available.

\section{Contributions}
Conceptualization: Ali Nazari, 
Data Curation: Ali Nazari, Bardiya Kariminia, 
Formal Analysis: Ali Nazari, Bardiya Kariminia,
Investigation: Ali Nazari, Bardiya Kariminia,
Methodology: Ali Nazari, Bardiya Kariminia,
Software: Ali Nazari, Bardiya Kariminia,
Validation: Ali Nazari,
Visualization: Ali Nazari,
Writing - Original Draft: Ali Nazari,
Writing - Review \& Editing: Ali Nazari, Bardiya Kariminia, Mohsen Ebrahimi Moghadam,
Project Administration: Mohsen Ebrahimi Moghadam,
Supervision: Mohsen Ebrahimi Moghadam

\section{Consent to Publish declaration}
Not applicable.
\section{Consent to Participate declaration}
Not applicable.
\section{Ethics declarations}
Not applicable.
\section{Funding}
Not applicable.
\section{Competing interests}
The authors declare no competing interests.

\begin{appendices}
	
	\section{Confusion matrices of the CUSP, SAM, and RA-GAN models}
	\label{sec:appendix}
	
	\begin{figure*}[htp]
		\centering
		\begin{tabular}{@{} m{5mm} @{} c@{} c@{}} 
			\textbf{age} & \textbf{RA-GAN} & \textbf{SAM-GAN} \\ \hline
			\rotatebox{90}{\hspace{-4cm}\textbf{20}} & 
			\raisebox{-\height}{\includegraphics[width=0.45\textwidth, trim={1.5cm 0.5cm 4.5cm 1cm},clip]{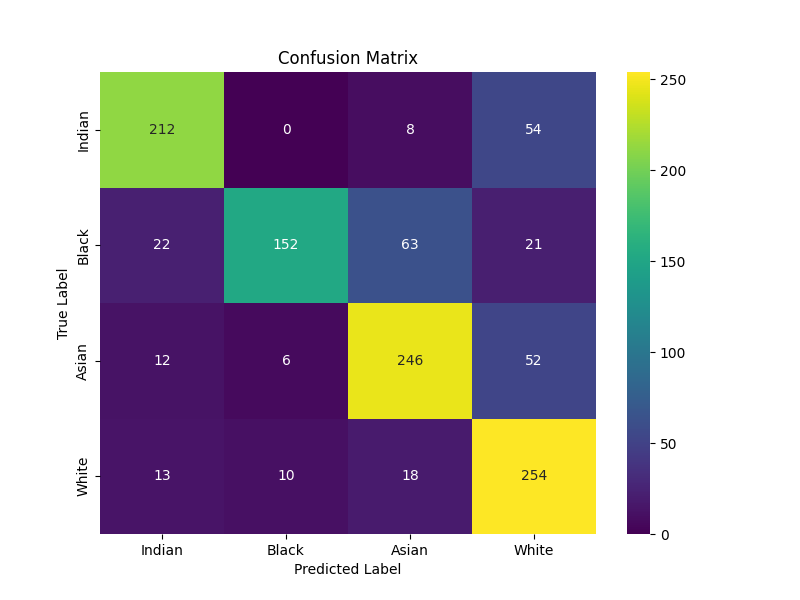}} &
			\raisebox{-\height}{\includegraphics[width=0.45\textwidth, trim={0.1cm 0.1cm 2.2cm 0.1cm},clip]{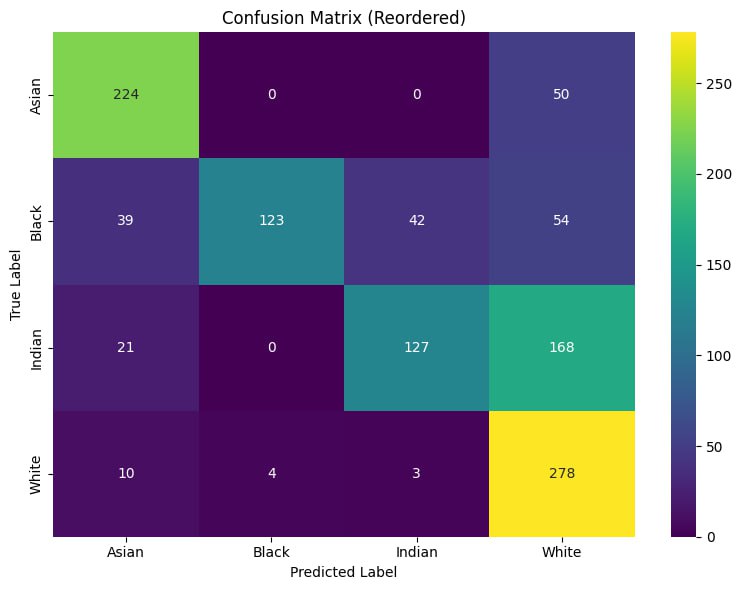}} \\
			
			\rotatebox{90}{\hspace{-4cm}\textbf{30}} & 
			\raisebox{-\height}{\includegraphics[width=0.45\textwidth, trim={1.5cm 0.5cm 4.5cm 1cm},clip]{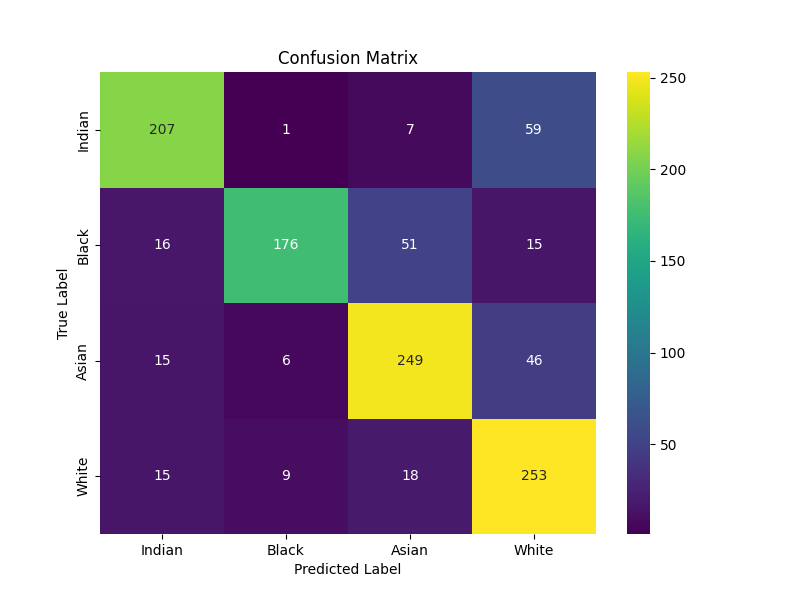}} &
			\raisebox{-\height}{\includegraphics[width=0.45\textwidth, trim={0.1cm 0.1cm 2.2cm 0.1cm},clip]{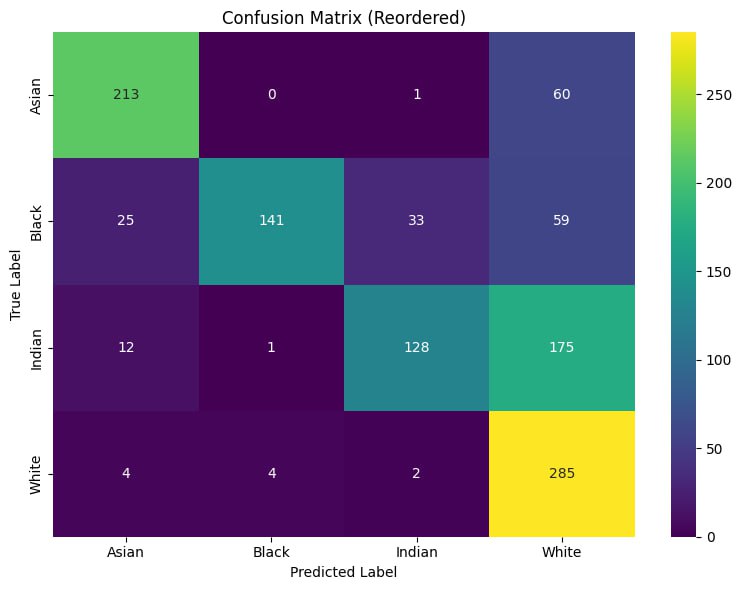}}
			\\
			
			\rotatebox{90}{\hspace{-4cm}\textbf{40}} & 
			\raisebox{-\height}{\includegraphics[width=0.45\textwidth, trim={1.5cm 0.5cm 4.5cm 1cm},clip]{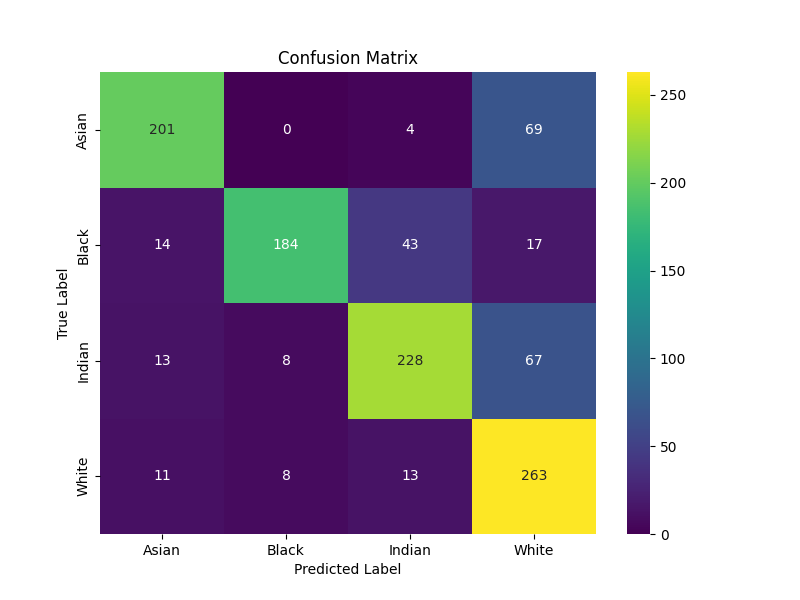}} &
			\raisebox{-\height}{\includegraphics[width=0.45\textwidth, trim={1.5cm 0.5cm 4.5cm 1cm},clip]{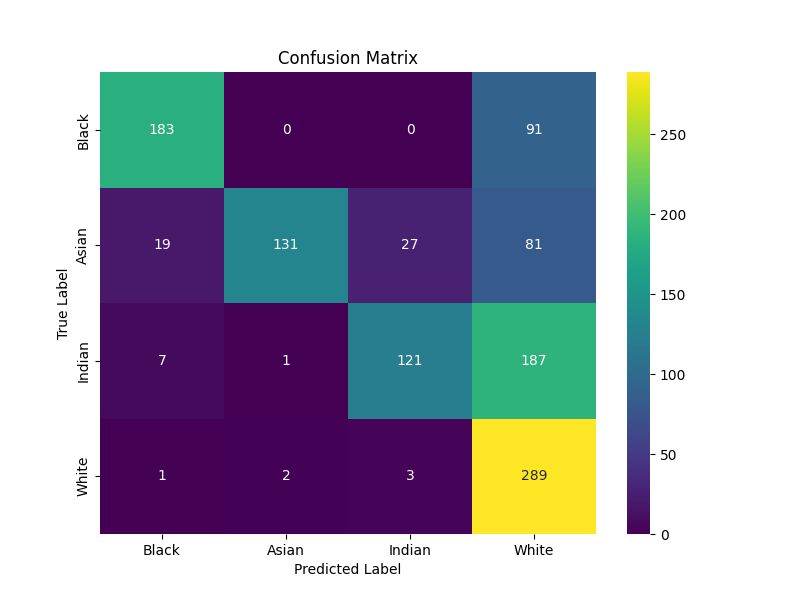}} \\        
		\end{tabular}
		\caption{Confusion matrices of age progression (20-40) for RA-GAN and SAM-GAN models.}
		\label{fig:race_confusion_matrices-1}
	\end{figure*}
	
	\begin{figure*}[htp]
		\centering
		\resizebox{0.7\textwidth}{!}{
			\begin{tabular}{@{} m{5mm} @{} c @{} c@{} @{}} 
				\textbf{age} & \textbf{RA-GAN} & \textbf{SAM-GAN} \\ \hline
				
				\rotatebox{90}{\hspace{-4cm}\textbf{50}} & 
				\raisebox{-\height}{\includegraphics[width=0.45\textwidth, trim={1.5cm 0.5cm 4.5cm 1cm},clip]{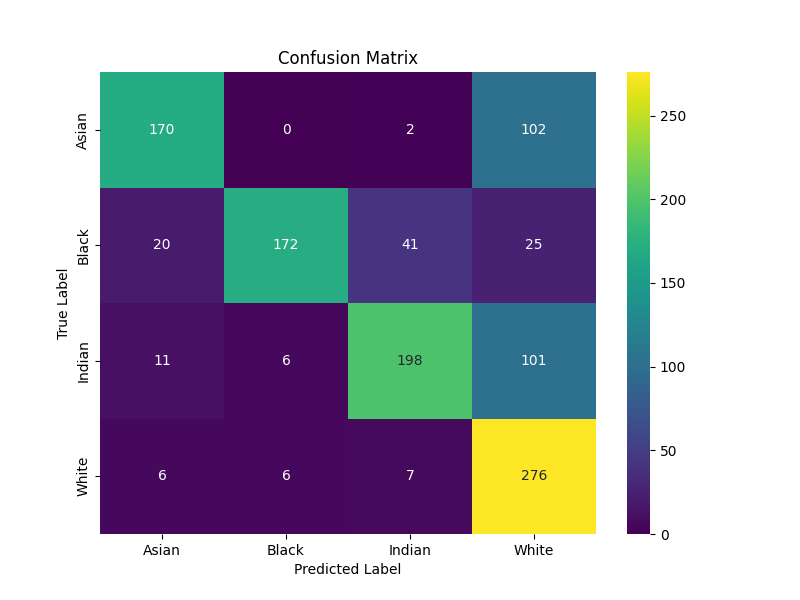}} &
				\raisebox{-\height}{\includegraphics[width=0.45\textwidth, trim={1.5cm 0.5cm 4.5cm 1cm},clip]{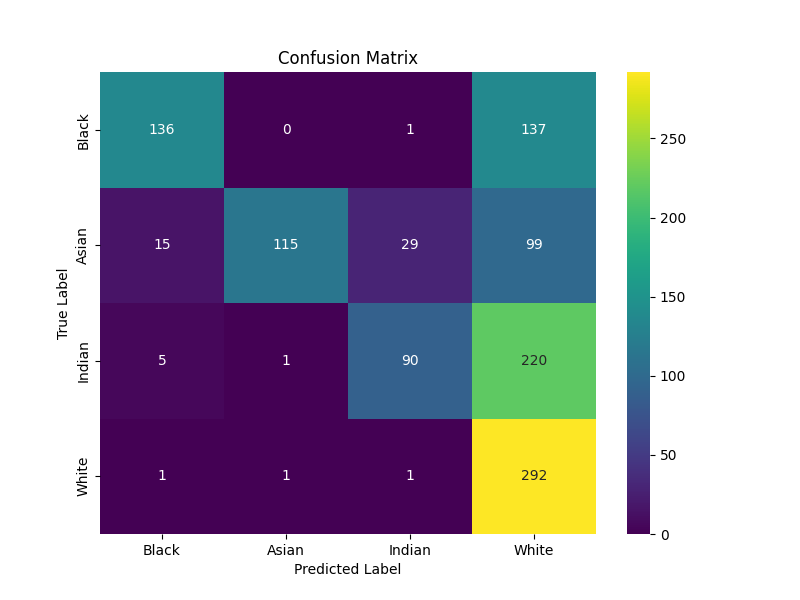}} \\
				
				\rotatebox{90}{\hspace{-4cm}\textbf{60}} & 
				\raisebox{-\height}{\includegraphics[width=0.45\textwidth, trim={1.5cm 0.5cm 4.5cm 1cm},clip]{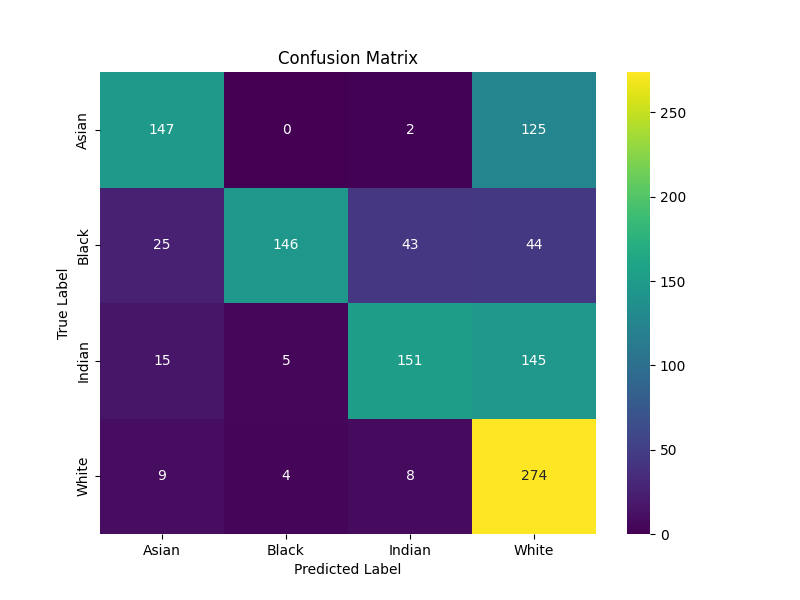}} &
				\raisebox{-\height}{\includegraphics[width=0.45\textwidth, trim={1.5cm 0.5cm 4.5cm 1cm},clip]{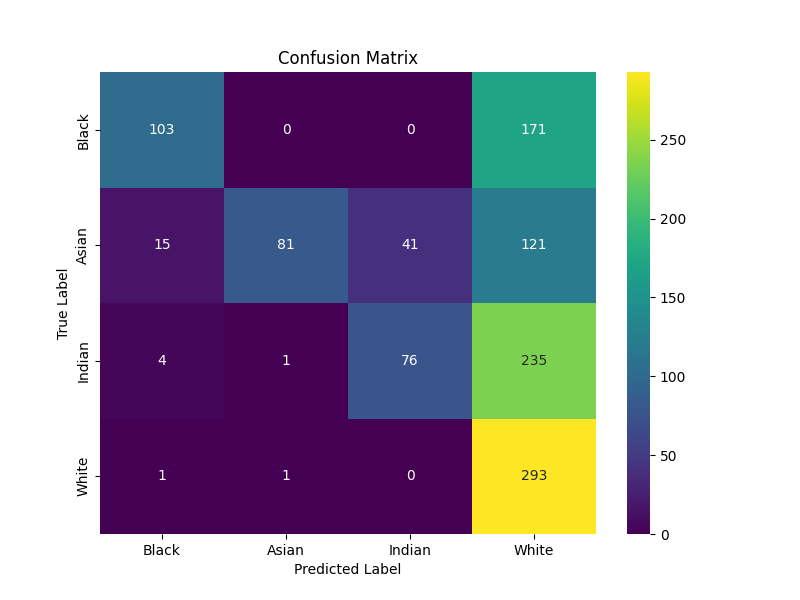}} \\
				
				\rotatebox{90}{\hspace{-4cm}\textbf{70}} & 
				\raisebox{-\height}{\includegraphics[width=0.45\textwidth, trim={1.5cm 0.5cm 4.5cm 1cm},clip]{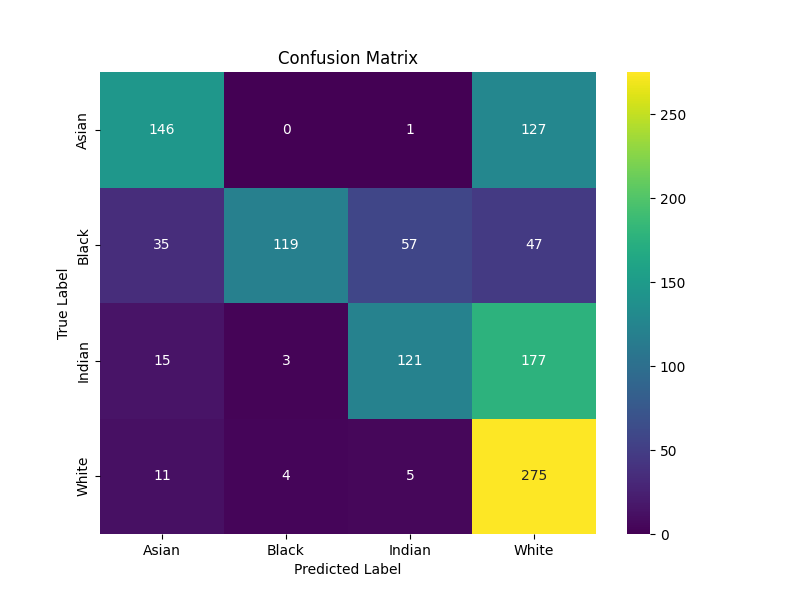}} &
				\raisebox{-\height}{\includegraphics[width=0.45\textwidth, trim={1.5cm 0.5cm 4.5cm 1cm},clip]{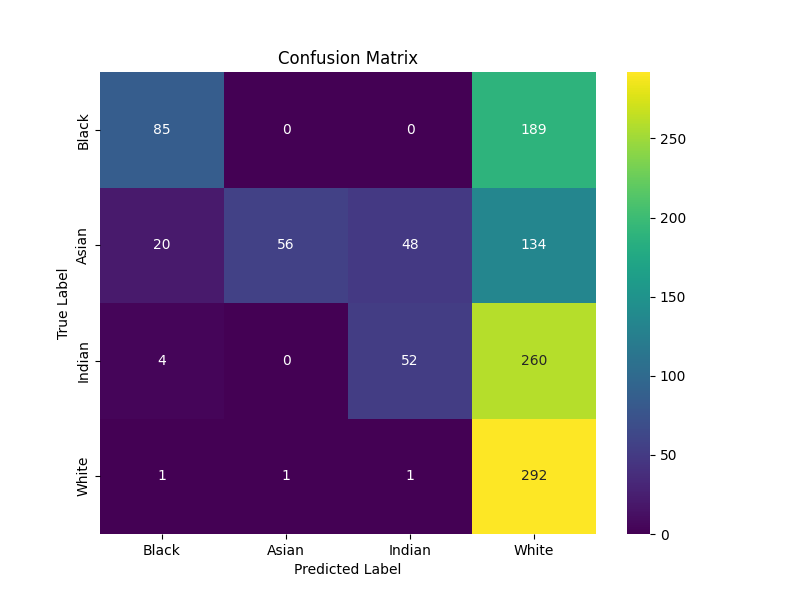}} \\
				
				\rotatebox{90}{\hspace{-4cm}\textbf{80}} & 
				\raisebox{-\height}{\includegraphics[width=0.45\textwidth, trim={1.5cm 0.5cm 4.5cm 1cm},clip]{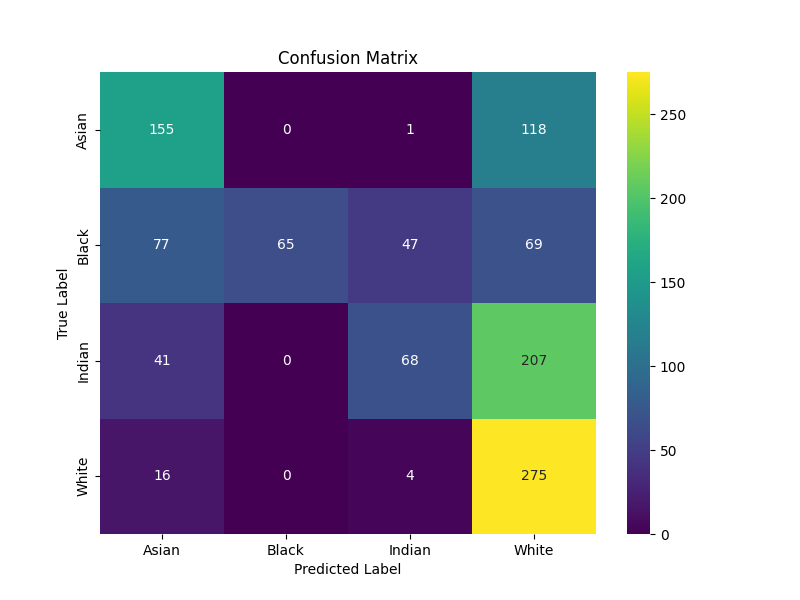}} &
				\raisebox{-\height}{\includegraphics[width=0.45\textwidth, trim={1.5cm 0.5cm 4.5cm 1cm},clip]{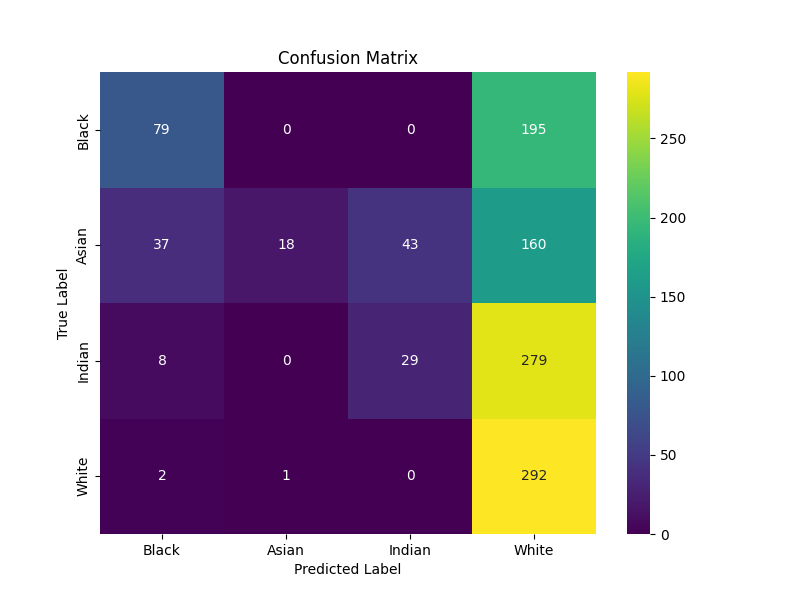}} \\
			\end{tabular}
		}	
		\caption{Confusion matrices of age progression (50-80) for RA-GAN and SAM-GAN models.}
		\label{fig:race_confusion_matrices-2}    
	\end{figure*}
	
	\begin{figure*}[htp]
		\centering
		\resizebox{0.9\textwidth}{!}{
			\begin{tabular}{@{} c @{} c@{} @{}} 
				
				{\textbf{20}} & {\textbf{30}} \\
				
				\includegraphics[width=0.5\textwidth, trim={1.5cm 0.5cm 2.9cm 1cm},clip]{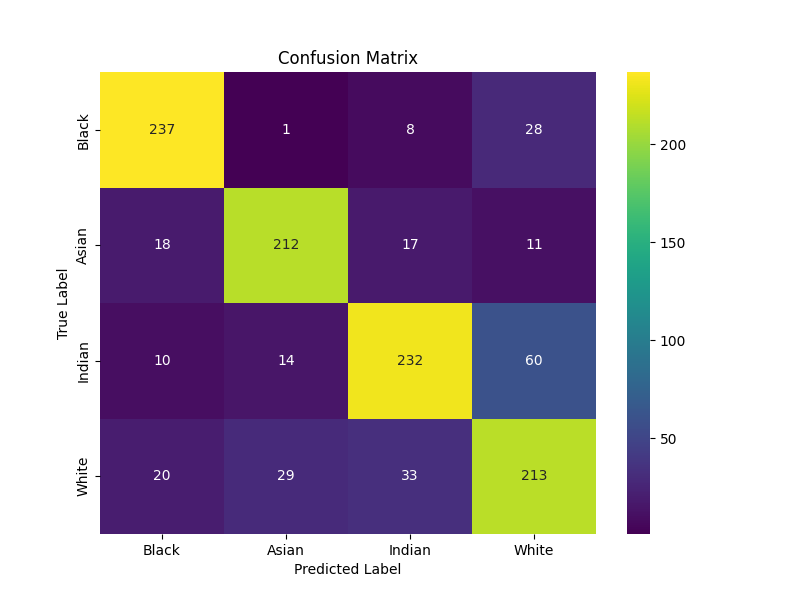} &
				\includegraphics[width=0.5\textwidth, trim={1.5cm 0.5cm 2.9cm 1cm},clip]{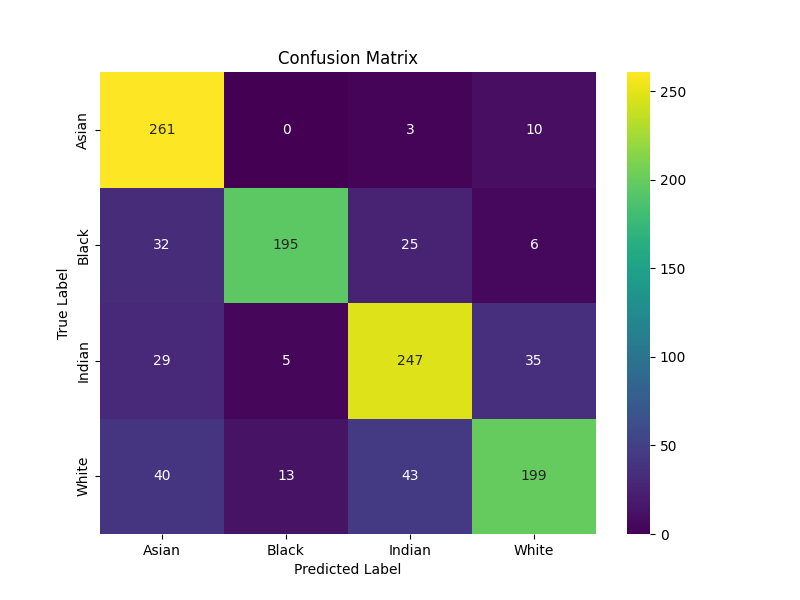} \\
				
				{\textbf{40}} & {\textbf{50}} \\
				
				\includegraphics[width=0.5\textwidth, trim={1.5cm 0.5cm 2.9cm 1cm},clip]{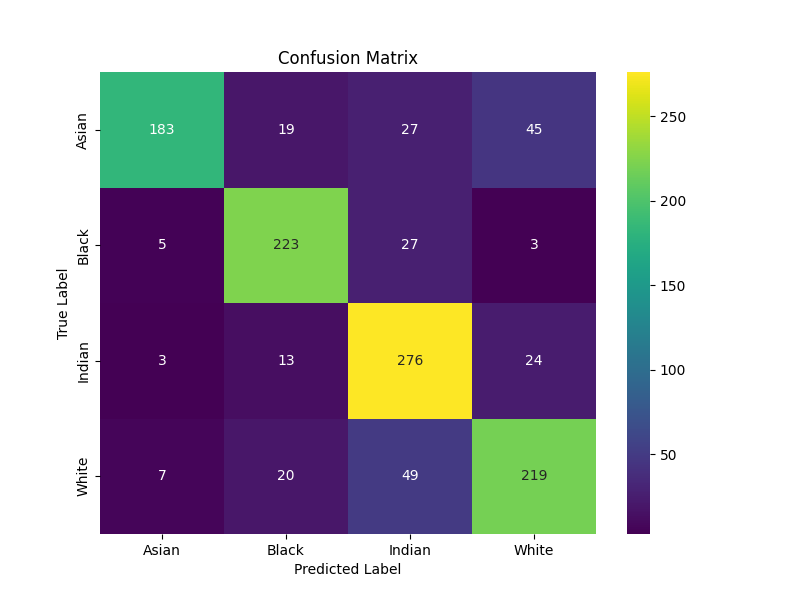} &
				\includegraphics[width=0.5\textwidth, trim={1.5cm 0.5cm 2.9cm 1cm},clip]{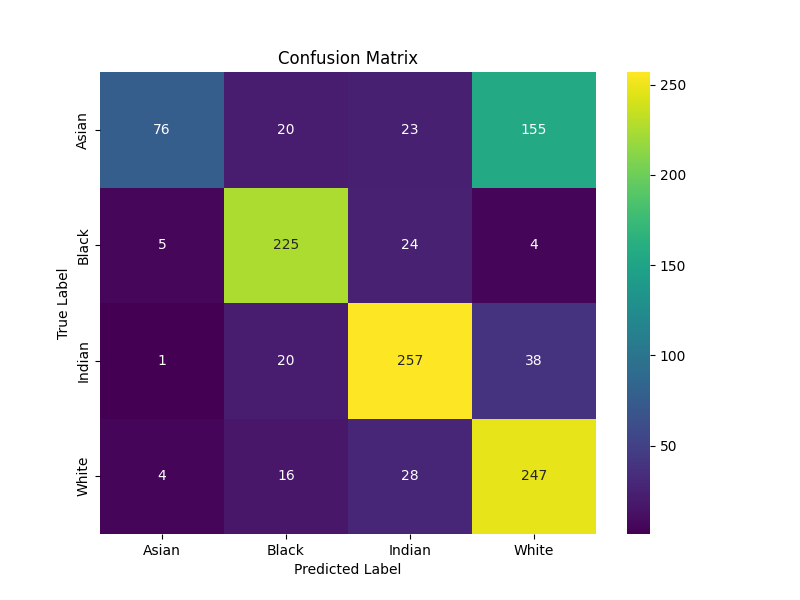} \\                
				
				\multicolumn{2}{c}{\textbf{60}}\\
				\multicolumn{2}{c}{\includegraphics[width=0.5\textwidth, trim={1.5cm 0.5cm 2.9cm 1cm},clip]{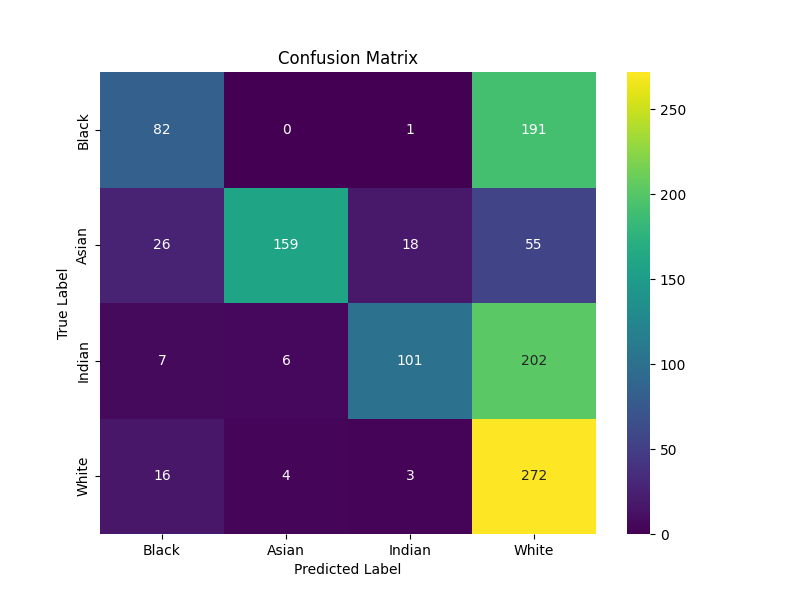}}\\
			\end{tabular}
		}
		\caption{The confusion matrices of age progression for the CUSP model. The numbers above the matrices are age groups.}
		\label{fig:race_confusion_matrices-cusp}
	\end{figure*}
	
\end{appendices}


\bibliography{sn-bibliography}

\end{document}